**Delineate Anything Flow: Fast, Country-Level Field Boundary Detection from Any Source**

*Mykola Lavreniuk[a*,b], Nataliia Kussul[b,c,d], Andrii Shelestov[b,d], Yevhenii Salii[b,d], Volodymyr Kuzin[b,d], Sergii Skakun[c], and Zoltan Szantoi[a,e]*

[a]European Space Agency, Frascati, 00044, Italy
[b]Space Research Institute NASU-SSAU, Kyiv, 03187, Ukraine
[c]University of Maryland, College Park, MD 20742, United States
[d]National Technical University of Ukraine “Igor Sikorsky Kyiv Polytechnic Institute”, Kyiv, 03056, Ukraine
[e]Stellenbosch University, Stellenbosch, South Africa

Corresponding author:
Mykola Lavreniuk
European Space Agency*, Frascati, 00044, Italy
nick_93@ukr.net

[*] International Research Fellow

**Highlights**

- DelAnyFlow: resolution-agnostic method for large-area field boundary mapping
- DelAny outperforms zero-shot SAM2 by up to 103% mAP and 400× faster inference
- Cross-tile unification reduces seam-induced field fragmentation by 4–6×
- FBIS-22M: largest benchmark with 673K images and 22.9M field instances (0.25–10 m)
- National-scale field boundaries map of Ukraine for 2024

**Abstract.**
Accurate satellite data-based field boundary delineation is essential for land management and agricultural statistics, but integrating tile-level predictions into continuous large-area vector products introduces topological errors and computational challenges. We present Delineate Anything Flow (DelAnyFlow), a framework for scalable field instance mapping that integrates the YOLOv11-based DelAny model with structured post-processing, cross-tile instance unification, vectorization, and topology-consistent simplification. DelAny is trained on the Field Boundary Instance Segmentation-22M (FBIS-22M) dataset, containing 672,909 multi-resolution patches from 0.25 to 10 m and 22.9 million field instances. It achieves the mean average precision with the intersection over union (IoU) threshold set at 0.5 (mAP@0.5) of 0.720 and mAP@0.5:0.95 of 0.477, while providing substantially faster inference than the zero-shot segment anything model (SAM2). Component-level evaluation shows that cross-tile unification reduces seam-induced field fragmentation by a factor of 4–6, while anchor-guided simplification reduces separated neighbouring fields from about 20% to 0.2% at comparable geometric compression. Demonstrating operational scalability, DelAnyFlow mapped Ukraine's entire 603,000 km$^2$ territory in 5.4 hours on a single workstation. Quantitative comparisons with independent operational products show larger IoU values and small-object recall (producer accuracy), particularly for fragmented agricultural landscapes; DelAnyFlow reaches an IoU of 0.971 and a small-object recall of 0.878, compared with 0.926 and 0.405 for the best operational product. The framework therefore provides a reproducible pathway from tile-level GeoAI predictions to spatially continuous, topologically consistent field-boundary products at regional and national scales. Code, models, the dataset, and national vectors are available at https://lavreniuk.github.io/Delineate-Anything/.

## 1. Introduction

Agricultural parcels (Land Parcel Identification System: LPIS) and fields link satellite data to land management, enabling crop mapping, phenology retrieval (Graesser and Ramankutty, 2017), field-level yield estimation (Kang and Özdoğan, 2019), and monitoring of cropland burning (Hall et al., 2021), war damage (Kussul et al., 2023), and land degradation (Kussul et al., 2017). Accurate boundaries are vital for policy; e.g., the European Union distributes over €45 billion annually under the Common Agricultural Policy (CAP) (Scown et al., 2020), using remote sensing to validate declarations and prioritize inspections (van der Velden et al., 2025). However, structural errors (under/over-segmentation, incomplete edges) heavily distort area estimates, meaning automated tools usually assist rather than replace expert validation in LPIS workflows (Erden et al., 2015).

Traditional edge detection and watershed algorithms suffered from intensive parameter tuning and poor scalability (Turker and Kok, 2013). Deep learning introduced semantic segmentation via U-Net (Garcia-Pedrero et al., 2019), later evolving into ResUNet-a (Diakogiannis et al., 2020), and U-Net integrated with Kolmogorov-Arnold Networks (Cambrin et al., 2024). Multi-task frameworks enhanced edges by jointly predicting boundaries, extent, and distance maps (Diakogiannis et al., 2020), such as FracTAL ResUNet, which fuses distance channels with hierarchical watershed post-processing (Waldner et al., 2021), and more recent edge-aware multi-task networks including SEANet (Li et al., 2023) and MEGRDNet (Yan et al., 2026). Yet, pixel-wise semantic approaches cause topological flaws (intersecting boundaries, merged fields), while hybrid graph or watershed corrections introduce heavy computational complexity without ensuring instance-level separation (Duvvuri and Kambhammettu, 2023; Wagner and Oppelt, 2020).

To address semantic limitations, research transitioned to instance-level modeling. Advanced computer vision architectures like Co-DETR (Zong et al., 2023), ViT-Adapter (Chen et al., 2023), EVP (Lavreniuk et al., 2025a), and real-time YOLO variants (Wang et al., 2024) provide superior efficiency in natural images (Lavreniuk and Lavreniuk, 2025c), but agricultural deployment is throttled by a lack of domain-specific, instance-annotated data.

Few studies have applied instance segmentation directly to field boundary delineation. Kerner et al. (2024b) investigated multi-region transfer learning under limited labels, while Cao et al. (2023) adapted Mask R-CNN with a dual-attention feature pyramid network for small and irregular parcels. These studies demonstrated the potential of instance-level representations but remained limited in geographic extent, spatial resolution, or sensor diversity. In our previous study (Lavreniuk et al., 2025b), we advanced this direction by introducing the DelAny instance-segmentation model and the multi-sensor, multi-resolution FBIS-22M dataset. That work focused on tile-level field delineation, model accuracy, computational efficiency, and cross-resolution applicability. Related foundation-model approaches, including SAM and SAM2 (Kirillov et al., 2023; Ravi et al., 2025), and adaptations based on prompt engineering (Osco et al., 2023), weak supervision (Sun et al., 2024), multi-scale attention for SAR (Han et al., 2023), and boundary-prior fusion (Long et al., 2024), also primarily address delineation at the image or tile level.

However, accurate tile-level delineation does not directly yield a continuous large-area geospatial product. Regional and national mapping requires thousands of independently processed tiles to be integrated into a single vector layer. Fields crossing tile boundaries may be duplicated, truncated, or represented by conflicting instances, while independent vectorization and simplification may disrupt shared boundaries and topology. Such scale-transition errors are not captured by conventional tile-level segmentation metrics.

Existing datasets and operational products also remain limited in scale, geographic coverage, resolution, or reproducibility. AI4SmallFarms (Persello et al., 2023) contains 62 images; AI4Boundaries (d'Andrimont et al., 2023) about 55,000 images; PASTIS (Garnot and Landrieu, 2021) about 124,000 objects; and Fields of the World (Kerner et al., 2024a) 70,484 10-m Sentinel-2 images. Recent VHR benchmarks such as FHAPD (Zhao et al., 2025) and GTPBD (Zhang et al., 2025) provide finer detail but remain geographically localized or focused on specific landscapes. Operational products are also difficult to reproduce: Sinergise Solutions does not disclose its training data, while the NASA Harvest Ukraine product relies on edge detection applied to PlanetScope time series rather than instance segmentation (Sadeh et al., 2025).

The central research question is therefore: **How can independent tile-level field delineations be integrated into a spatially continuous and topologically consistent country-scale vector map?** We hypothesize that **cross-tile field unification combined with topology-consistent vector simplification can reduce tile-boundary inconsistencies while preserving field geometry and delineation accuracy.**

To test this hypothesis, we extend DelAny and FBIS-22M into Delineate Anything Flow (DelAnyFlow), a large-area geospatial mapping framework integrating adaptive tiling, structured post-processing, cross-tile instance unification, and topology-consistent vector simplification. Our main contribution is the transition from tile-level GeoAI predictions to a continuous large-area vector product and its quantitative evaluation at component, regional, and national scales, including country-scale deployment over Ukraine (603,000 km²) and comparison with independent operational field-boundary products.

The specific contributions are: (1) DelAnyFlow, a framework that converts independent tile-level instance predictions into a continuous country-scale vector map through cross-tile instance unification and anchor-guided, topology-consistent simplification; (2) a quantitative evaluation of this transition at three scales: a component-level ablation of tile-boundary duplication, fragmentation and topological consistency, regional experiments on spatial resolution, acquisition date and spectral input, and a strict spatial holdout of the benchmark; (3) a country-scale deployment over Ukraine (603,000 km², 5.4 hours on a single desktop) with a comparison against three operational field-boundary products, with code, models and data openly released.

## 2. Methods and Data

In this study, we propose the Delineate Anything Flow (DelAnyFlow) framework, a resolution-agnostic methodology designed for large-scale, automated agricultural field boundary extraction. We define resolution-agnostic as the capacity of a single operational pipeline to ingest and process multi-sensor imagery from 0.25 to 10 m without requiring architectural redesign, retraining, or resolution-specific parameter tuning. By reformulating boundary delineation from a semantic pixel classification task into an instance segmentation problem, DelAnyFlow mitigates edge misalignments and prevents the artificial merging of adjacent parcels. The workflow couples a custom-trained deep learning model (DelAny) with an automated post-processing pipeline consisting of adaptive tiling, hierarchical morphological refinement, cross-tile vector unification, and geometric simplification (Figure 1).

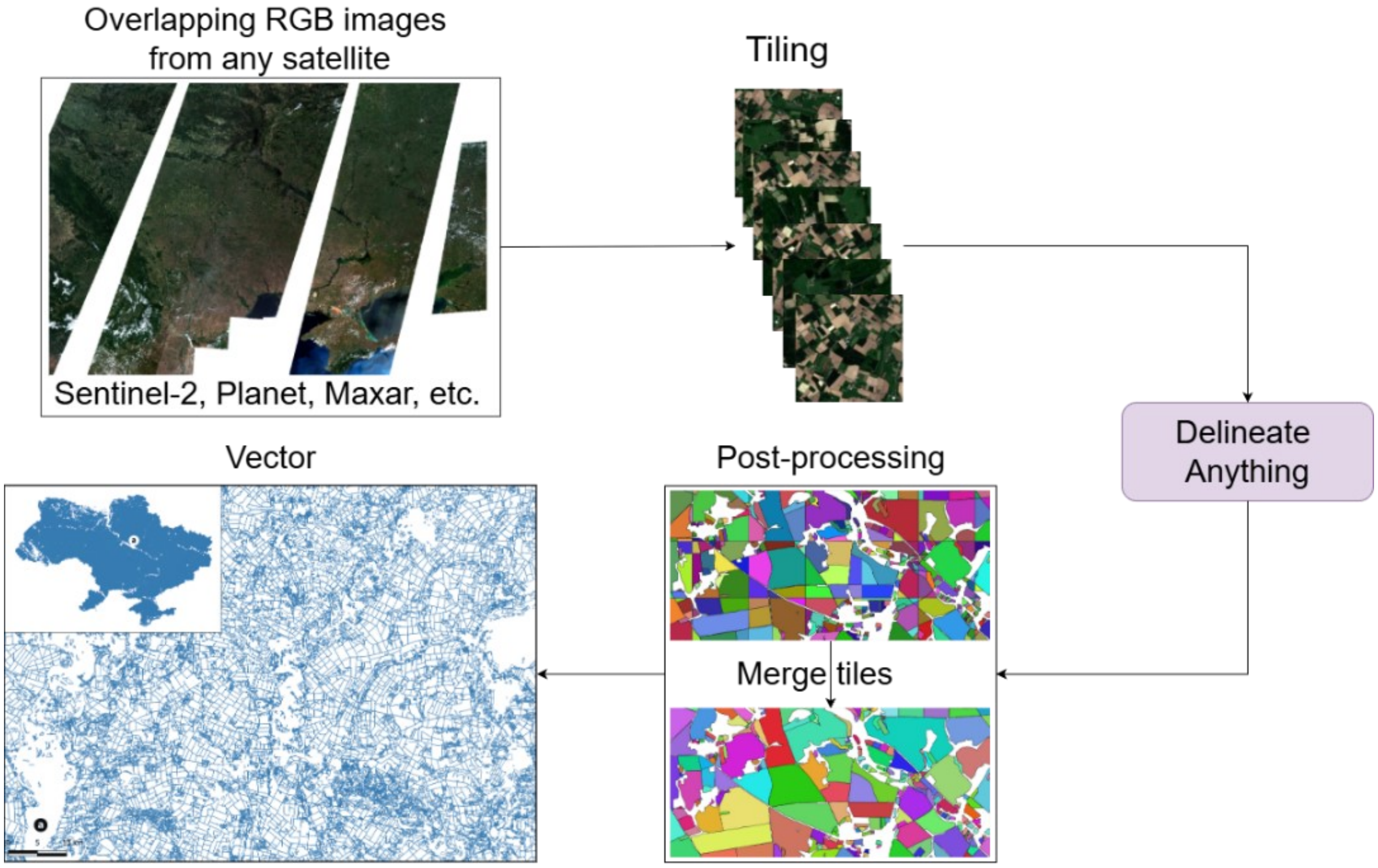

Figure 1. General workflow of the proposed DelAnyFlow methodology for national-scale field instance segmentation and boundary vectorization from multi-resolution satellite imagery.

### 2.1. Reframing Field Boundary Delineation as Instance Segmentation

Traditional semantic segmentation models for field edge detection are evaluated using boundary Intersection over Union (IoU), which is computed exclusively on the outermost boundary pixels. This metric is hyper-sensitive to minor spatial misalignments: a displacement of a few pixels can reduce the boundary IoU to 0.08, even when the predicted geometry precisely tracks the ground truth (Figure 2b). Conversely, instance IoU accounts for all pixels within the field body, yielding a stable score of 0.98 under identical conditions (Figure 2e) by prioritizing global object delineation over pixel-level edge matching. Both metrics share the general IoU algebraic structure but operate on distinct spatial domains:

$$IoU_{bnd} = \frac{|B(P) \cap B(G)|}{|B(P) \cup B(G)|},\ IoU_{inst} = \frac{|P \cap G|}{|P \cup G|},$$

where $B(\cdot)$ defines the boundary edge pixels of a polygon, while $P$ and $G$ denote the complete predicted and reference ground truth field regions, respectively.

Crucially, boundary IoU fails to adequately penalize under-segmentation errors where adjacent fields are merged. A partially omitted edge segment can still retain a boundary IoU of 0.93 (Figure 2c) despite failing to separate distinct parcels. Instance IoU correctly registers this structural error, dropping to 0.54 (Figure 2f). Preserving individual field boundaries is mandatory for field-level crop classification, yield modeling, and agricultural inventory. Reformulating the task as instance segmentation forces the model to generate closed, independent masks per field, ensuring geometric and topological consistency during large-scale operational mapping.

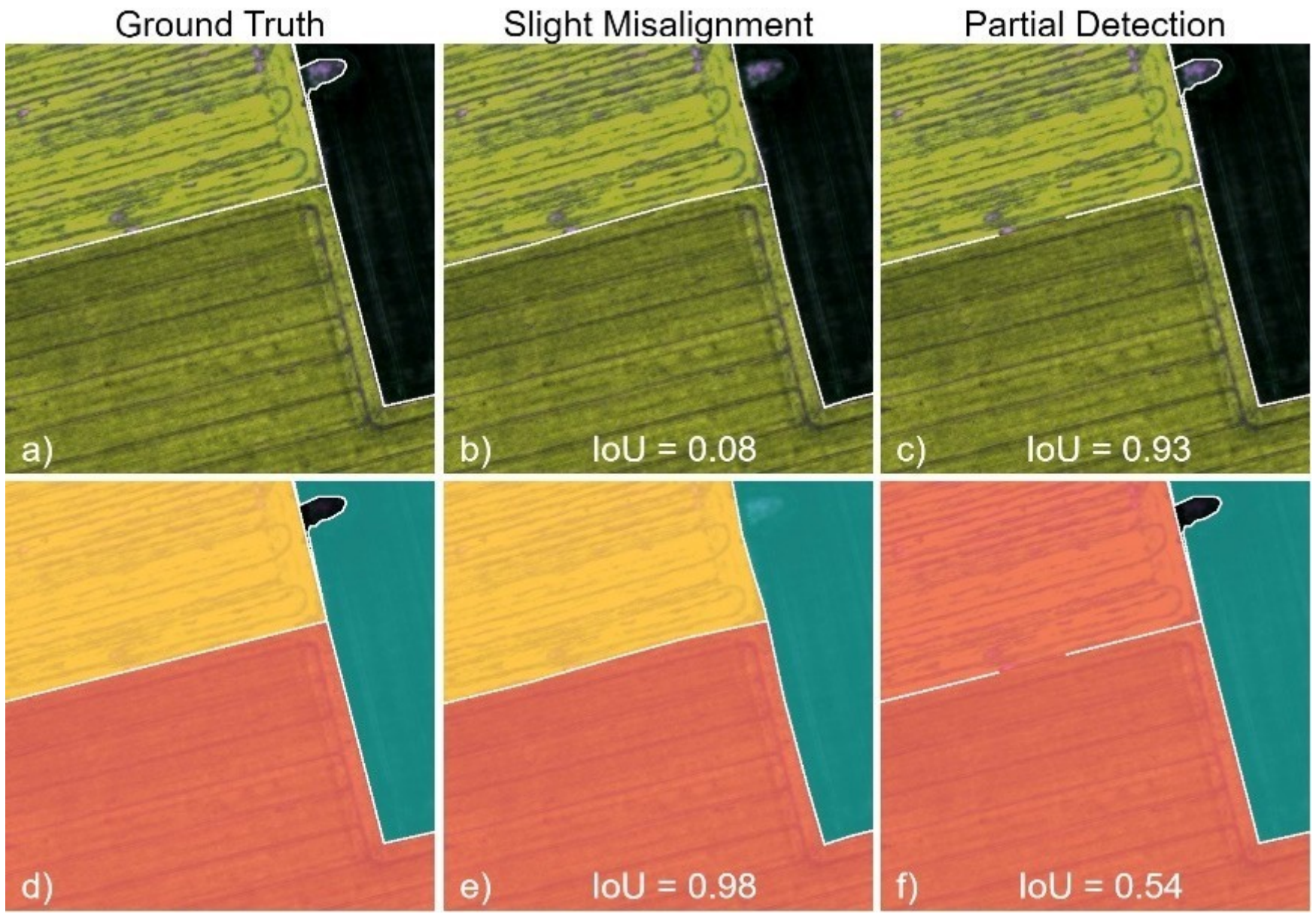

Figure 2. Comparison of task formulations and evaluation metrics for field boundary delineation: (a, d) reference ground truth; (b) low boundary IoU caused by minor edge misalignment; (c) high boundary IoU failing to capture field merging errors; (e) robust instance IoU under identical edge misalignment; (f) reduced instance IoU correctly penalizing the merging of adjacent fields.

### 2.2. Field Boundary Instance Segmentation Dataset

While computer vision models leverage massive general-purpose image datasets like LAION-5B (Schuhmann et al., 2022) and SA-1B (Kirillov et al., 2023), specialized remote sensing datasets for field delineation remain localized and small. To resolve this data gap, we developed the Field Boundary Instance Segmentation-22M (FBIS-22M) dataset. Containing 672,909 high-resolution satellite patches and 22,926,427 individual field masks, FBIS-22M is over twelve times larger than AI4Boundaries (d'Andrimont et al., 2023), the previous largest open-source benchmark (Table 1). This footprint substantially exceeds existing agricultural datasets in geographic coverage, resolution diversity, and instance count, establishing a new benchmark for automated land parcel identification. Furthermore, FBIS-22M is the first dataset of its kind to incorporate high-resolution commercial satellite imagery alongside public data, integrating Sentinel-2 (https://www.esa.int/Applications/Observing_the_Earth/Copernicus/Sentinel-2), PlanetScope (https://earth.esa.int/eogateway/missions/planetscope), Pleiades (https://earth.esa.int/eogateway/missions/pleiades), and very high-resolution commercial satellites operated by Maxar (now Vantor, https://vantor.com/product/worldview/), providing diverse heterogeneity across sensor modalities.

Table 1. Comparison of the FBIS-22M dataset with existing general computer vision and agricultural field delineation benchmarks.

| Dataset | Resolution | # Images | # Instances |
|---|---|---|---|
| General Computer Vision Datasets | | | |
| LAION-5B (Schuhmann et al., 2022) | - | 5.85B | - |
| COCO (Lin et al., 2014) | - | 330K | 1.5M |
| Open Images (Kuznetsova et al., 2020) | - | 998K | 2.8M |
| SA-1B (Kuznetsova et al., 2020) | - | 11M | 1.1B |
| Field Boundary Delineation Datasets | | | |
| Farm Parcel (Aung et al., 2020) | 10 m | 2K | - |
| India10K (Wang et al., 2022) | - | - | 10K |
| PASTIS (Garnot and Landrieu, 2021) | 10 m | 2K | 124K |
| PASTIS-R (Garnot et al., 2022) | 10 m | 2K | 124K |
| PASTIS-HD | 1 and 10 m | 2K | 124K |
| AI4SmallFarms (Persello et al., 2023) | 10 m | 62 | 439K |
| AI4Boundaries (d'Andrimont et al., 2023) | 1 and 10 m | 55K | 2.5M |
| Fields of The World (Kerner et al., 2024a) | 10 m | 70K | 1.63M |
| FBIS-22M | 0.25 to 10 m | 673K | 22.9M |

FBIS-22M integrates imagery across a wide resolution spectrum (0.25, 0.3, 0.5, 1, 1.2, 2, 3, and 10 m), ingesting public Sentinel-2 data alongside commercial imagery from PlanetScope, Pleiades, and Maxar/Vantor WorldView constellations. The dataset spans diverse agricultural topographies across Austria, France, Luxembourg, the Netherlands, Slovakia, Slovenia, Spain, Sweden, and Ukraine (Figure 3), capturing structural densities from fewer than 10 to over 300 fields per image tile (Figure 4).

The dataset comprises three components. The first consists of manually filtered Sentinel-2 and aerial orthophoto tiles sourced from the AI4Boundaries benchmark. The second incorporates official LPIS registries from France and Slovakia; for these regions, ground-truth parcels were paired with corresponding Sentinel-2 and Planet imagery. The third component includes manual field delineations produced according to standard LPIS specifications on very-high-resolution Maxar basemaps, within the 2024 LPIS Pilot project for Ukraine (supported by Maxar, Pleiades, and Planet data).

All extracted tiles underwent quality control conducted by specialists in LPIS production. Tiles exhibiting heavy cloud/shadow cover, sensor artifacts, or spatial or topical misalignments between imagery and vector boundaries were removed. The final dataset is partitioned into 636,784 training and 36,125 test tiles. The test split was designed to prevent spatial leakage: 4,544 AI4Boundaries tiles across Austria (1,021), Spain (654), France (1,151), Luxembourg (81), the Netherlands (675), Sweden (708), and Slovenia (254), alongside 15,554 tiles from two dedicated holdout regions in Ukraine, all having zero spatial intersection with the training data. The remaining 16,024 tiles in Slovakia feature Planet data strictly disjoint from training Planet scenes, while Sentinel-2 tiles may share spatial footprints with training areas across sensors but are acquired on strictly different dates.

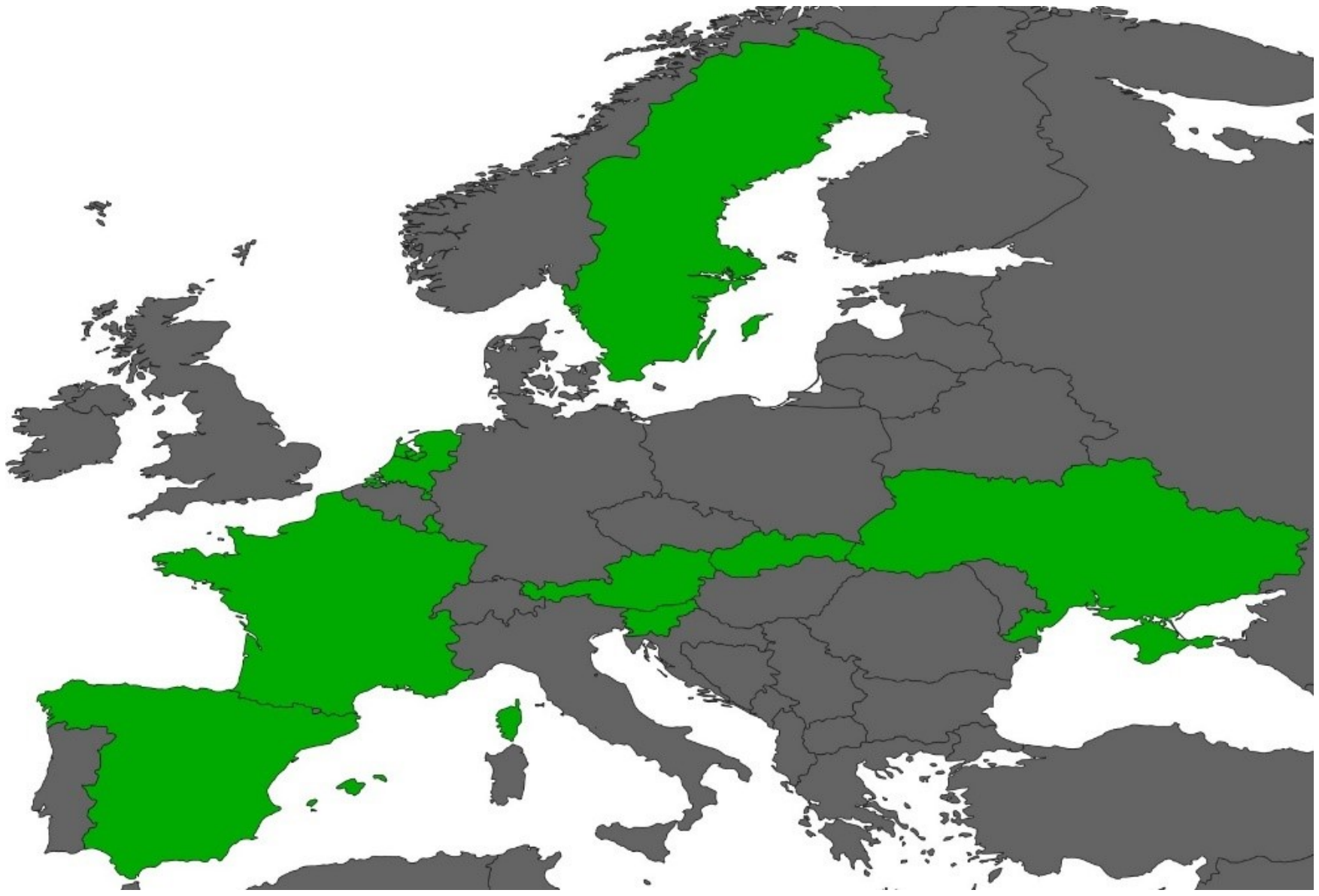

Figure 3. Geographic distribution of countries (highlighted in green) represented in the FBIS-22M dataset.

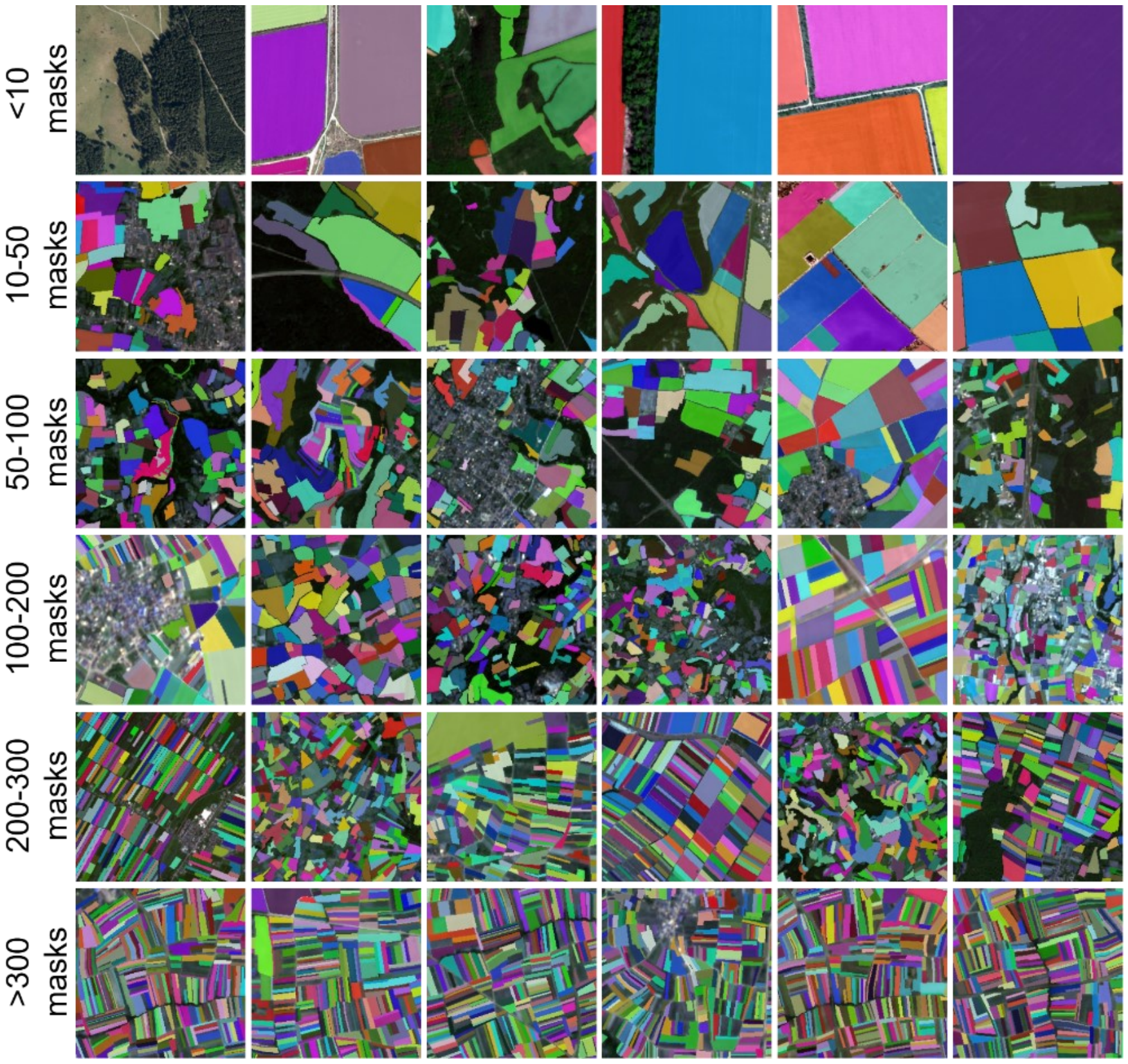

Figure 4. Examples of field boundary instance segmentation from the FBIS-22M dataset across varying structural densities and spatial resolutions. Each color represents a unique field instance.

## 2.3. Instance-Based Field Segmentation with Delineate Anything

DelAny follows the standard YOLOv11 backbone-neck-head design with an added instance-segmentation mask head; full architecture details are provided in Lavreniuk et al. (2025b) and the released code. DelAny ingests standard three-band optical inputs using the spectral channels closest to red, green, and blue wavelengths (e.g., Sentinel-2 bands B4, B3, and B2), ensuring direct cross-sensor compatibility without input reconfiguration. Ablation experiments show that the model can perform delineation with similar accuracy on false-color imagery, such as composites of different NIR timestamps or synthetic imagery obtained from time-series harmonics (Table 6).

The framework evaluates two structural variants: DelAny – 1.5× width multiplier, maximum 512 channels, 379 layers, and 62M parameters, and DelAny-S – 0.5× depth, 0.25× width, 203 layers, and 2.9M parameters. Both variations are trained natively on FBIS-22M under identical hyperparameters and operate zero-shot at inference without localized fine-tuning. The models process 512×512-pixel tiles extracted from normalized surface reflectance scenes via mapping 1–99 percentile of selected scene channels to 0–255 integers, outputting raw instance masks and localized bounding boxes directly to the post-processing engine.

### 2.4. DelAnyFlow Architectural Summary

The complete end-to-end DelAnyFlow pipeline coordinates data ingestion and vector delivery across five automated stages (Figure 5):

1. **Adaptive data preparation.** Resolution-aware tiling of multi-sensor inputs using spatial grids with predefined pixel overlaps.
2. **Instance segmentation.** Each tile is processed with the DelAny instance segmentation model (Section 2.3), which predicts field-level instances.
3. **Structured post-processing.** Predicted masks are refined using geometric and morphological operations, including mask cleaning and area-based filtering, to remove artifacts and improve boundary definition.
4. **Cross-tile field unification.** Overlapping field instances from adjacent tiles are merged using spatial overlap and geometric criteria to ensure topologically consistent geometries.
5. **Vectorization and simplification**. Extraction of raw contours and vertex reduction via anchor point frequency maps.

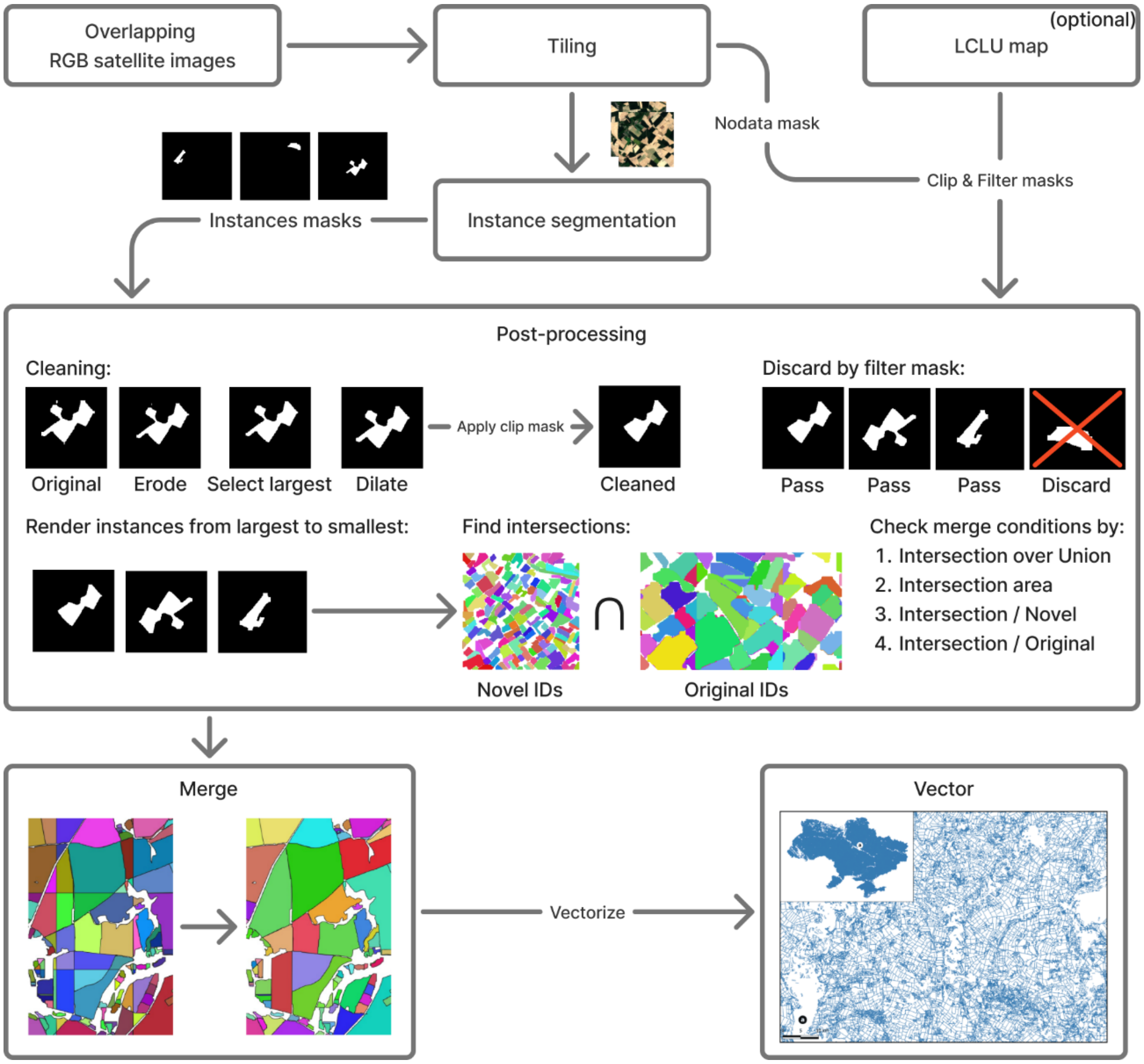

Figure 5. End-to-end workflow of the DelAnyFlow for large-scale field boundary delineation.

## 2.5. Output Integration and Final Boundary Generation

To convert independent tile-level raster outputs into a continuous, intersectionless vector database, DelAnyFlow applies a sequence of spatial, morphological, and geometric operators. Cropland boundaries are enforced through dual-level spatial masking. An optional quality mask excludes corrupted pixels, cloud artifacts, and static non-agricultural classes (e.g., water bodies, urban infrastructure) using available local land cover layers. A context mask is then generated by applying a spatial buffer to these exclusion zones and linear features like roads and forest boundaries. This flags transition areas susceptible to edge degradation, preventing the model from expanding field boundaries into non-cropland environments. Extracted masks are processed in descending order of geometric area to prioritize high-confidence interior field structures. Each instance undergoes hierarchical morphological cleaning to rectify segmentation noise:

1. **Erosion:** Cleaves thin raster bridges that incorrectly fuse separate adjacent fields.
2. **Connected Component Analysis:** Isolates the largest contiguous pixel cluster, discarding isolated fragment artifacts.
3. **Dilation:** Restores the valid field core back to its original geometric margin while maintaining corrected topological separation.

Following refinement, instances are checked against valid-pixel ratios and minimum size thresholds; sub-threshold masks are discarded.

To resolve redundant or conflicting field masks caused by overlapping tile boundaries, overlapping instances are evaluated via an interactive intersection test. Polygons meeting empirically derived IoU and area-proportional thresholds are merged into a unified geometric instance, preserving large continuous field extensions while maintaining fine-scale boundary definitions. The unified raster mosaic is vectorized into raw polygons, which undergo systematic topological validation to eliminate sliver artifacts and overlapping nodes. The final step compresses the dense vertex structures to ensure data scalability (Figure 6). Polygon margins are rasterized to build a vertex occurrence frequency map. Any vertex of the densified polygon with a preceding or succeeding vertex of strictly lower frequency is marked as a permanent anchor point. Intermediate line segments connecting these anchor points are then simplified using the Ramer-Douglas-Peucker (RDP) algorithm. This vertex reduction preserves characteristic field contours and sharp corners while generating a non-overlapping, topologically closed vector dataset optimized for regional and national agricultural inventory.

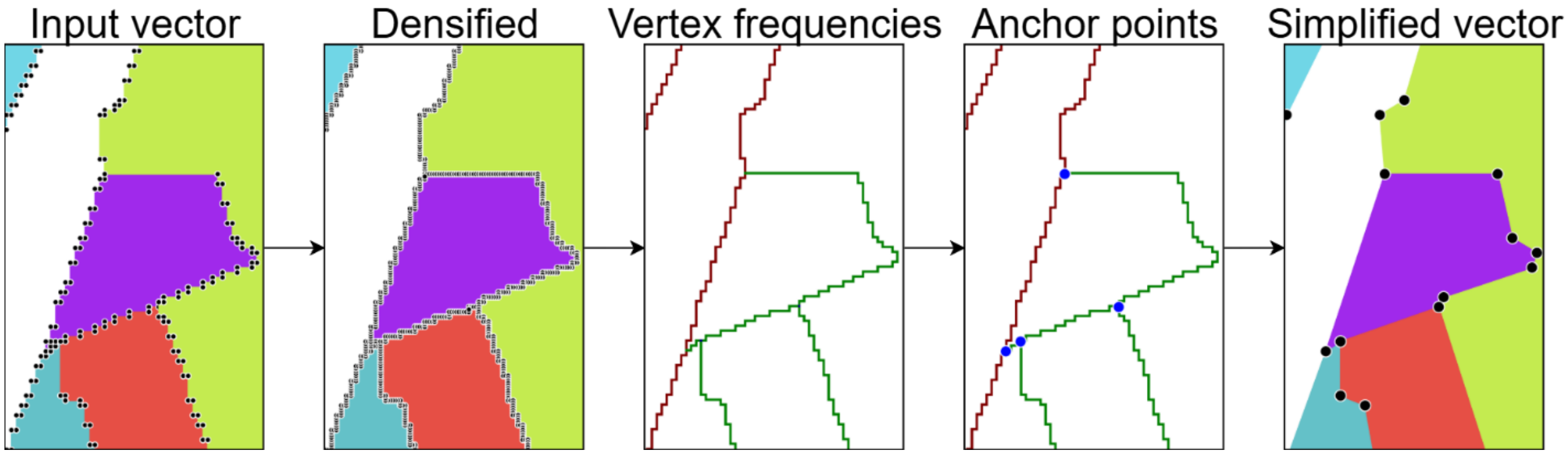

Figure 6. Overview of the proposed field boundary simplification method: raw vertex-dense polygons (left), anchor point detection and Ramer–Douglas–Peucker simplification, producing a topologically consistent vector output (right).

## 3. Results

To comprehensively evaluate the DelAny model and the DelAnyFlow framework, we executed six systematic experiments addressing core operational dimensions of automated field boundary delineation: (1) architectural benchmarking and model configuration, (2) zero-shot cross-region geographical feasibility, (3) training dataset scale and diversity dynamics, (4) input spatial resolution sensitivity, (5) multi-temporal phenological stability, and (6) spectral-temporal compositing and preprocessing strategies.

### 3.1. Training Configuration and Evaluation Metrics

Model validation was conducted using object-based and instance-level metrics. We report the mean Average Precision at a single Intersection over Union threshold (mAP@0.5) and averaged across ten progressive thresholds (mAP@[0.5:0.95], spanning 0.5 to 0.95 with 0.05 increments) adhering to the Microsoft COCO evaluation protocol (Lin et al., 2014):

$$AP = \sum_{i=1}^{m-1} [R(i) - R(i-1)] \times max(P(i), P(i-1)),$$

where $R(i)$ and $P(i)$ are the recall and precision after considering the $i$-th prediction in the ranked list of $m$ predictions sorted by confidence.

We additionally estimated IoU between the predicted and ground-truth cropland masks (each defined as the union of all polygons), and evaluated the recall rate for small objects, defined as the percentage of polygons below the 25th percentile in area covered by the prediction. To assess over- and under- segmentation errors, we evaluated the means of object-based indices OC (Over Classification), UC (Under Classification), and TC (Total Over- and Under Classification) (Long et al., 2022).

For the mask-level evaluation, where *P* is the predicted mask, and *G* is the reference ground-truth mask, the metric is defined as:

$$IoU = \frac{|P \cap G|}{|P \cup G|}$$

For object-based indices, where *G* is the reference shape of the object, and *P* is the predicted shape of the object yielding the largest IoU with the ground truth:

$$OC(G) = 1 - \frac{|G \cap P|}{|G|}, UC(G) = 1 - \frac{|G \cap P|}{|P|}, TC(G) = \sqrt{\frac{OC(G)^2 + UC(G)^2}{2}}$$

All models were trained using standardized 512 × 512 RGB image patches. Both models were trained for 30 epochs with a batch size of 320 (40 images per GPU) and a learning rate of $2\times10^{-5}$, using the standard YOLO loss function (Wang et al., 2024) with AdamW optimization, weight decay, and cosine decay with warmup, initialized from COCO pre-trained weights. Training used 8 NVIDIA H100 GPUs with FP16 mixed precision. Augmentations included horizontal/vertical flipping, color jittering, mixup, copy-paste, and mosaic augmentation (first 20 epochs only). Parameters used in DelAnyFlow are specified in Supplementary Material.

For SAM and SAM2, we used automatic mask generation ("segment everything") with the models' default grid-prompt sampling and standard confidence/stability-score thresholds, the only zero-shot configuration usable at operational scale, since field-level prompts or labels are not available a priori.

## 3.2. Performance Metrics and Benchmarking

Table 2 summarizes the performance metrics of the two modifications of Delineate Anything model and compares them with several state-of-the-art baselines, including MultiTLF (Kerner et al., 2023), SAM (Kirillov et al., 2023), and SAM2 (Ravi et al., 2025). Of these, MultiTLF was re-trained on the FBIS-22M dataset to ensure a consistent evaluation framework.

Table 2. Quantitative performance comparison of field boundary delineation models on the FBIS-22M test set. † indicates the model re-trained on the FBIS-22M dataset for consistency. Latency (ms) denotes the total inference time required to generate field boundaries for a 512 × 512-pixel patch on an NVIDIA A100 GPU. Best results are shown in bold.

| Method | mAP@0.5 | mAP@0.5:0.95 | Latency (ms) |
|---|---|---|---|
| MultiTLF † | 0.257 | 0.110 | 55.8 |
| SAM | 0.339 | 0.197 | 13605 |
| SAM2 | 0.382 | 0.235 | 10370 |
| **DelAny-S** | 0.632 | 0.383 | **16.8** |
| **DelAny** | **0.720** | **0.477** | 25.0 |

Both DelAny variants outperform all baselines across the evaluated metrics. DelAny surpasses SAM2 by 88.5% in mAP@0.5 and by 103% in mAP@[0.5:0.95]. The lightweight variant, DelAny-S, remains highly competitive despite its reduced parameter count, outperforming SAM2 by 65.5% and 63% across the same metrics. This accuracy advantage is paired with computational benefits: DelAny and DelAny-S achieve 415× and 617× lower average inference latency than SAM2, reflecting the efficiency gains of a purpose-built architecture over general-purpose segmentation foundation models.

A qualitative comparison (Figure 7) clarifies the source of this performance gap. MultiTLF performs adequately in landscapes with large, spectrally homogeneous fields, but its performance deteriorates in dense smallholder settings, where it tends to merge or omit adjacent plots. Conversely, SAM and SAM2 over-merge neighboring fields and frequently misclassify non-agricultural features (e.g., water bodies, grassland, and forest) as field instances. DelAny avoids both failure modes, maintaining accuracy across diverse field configurations and the full range of field sizes due to its instance-segmentation formulation.

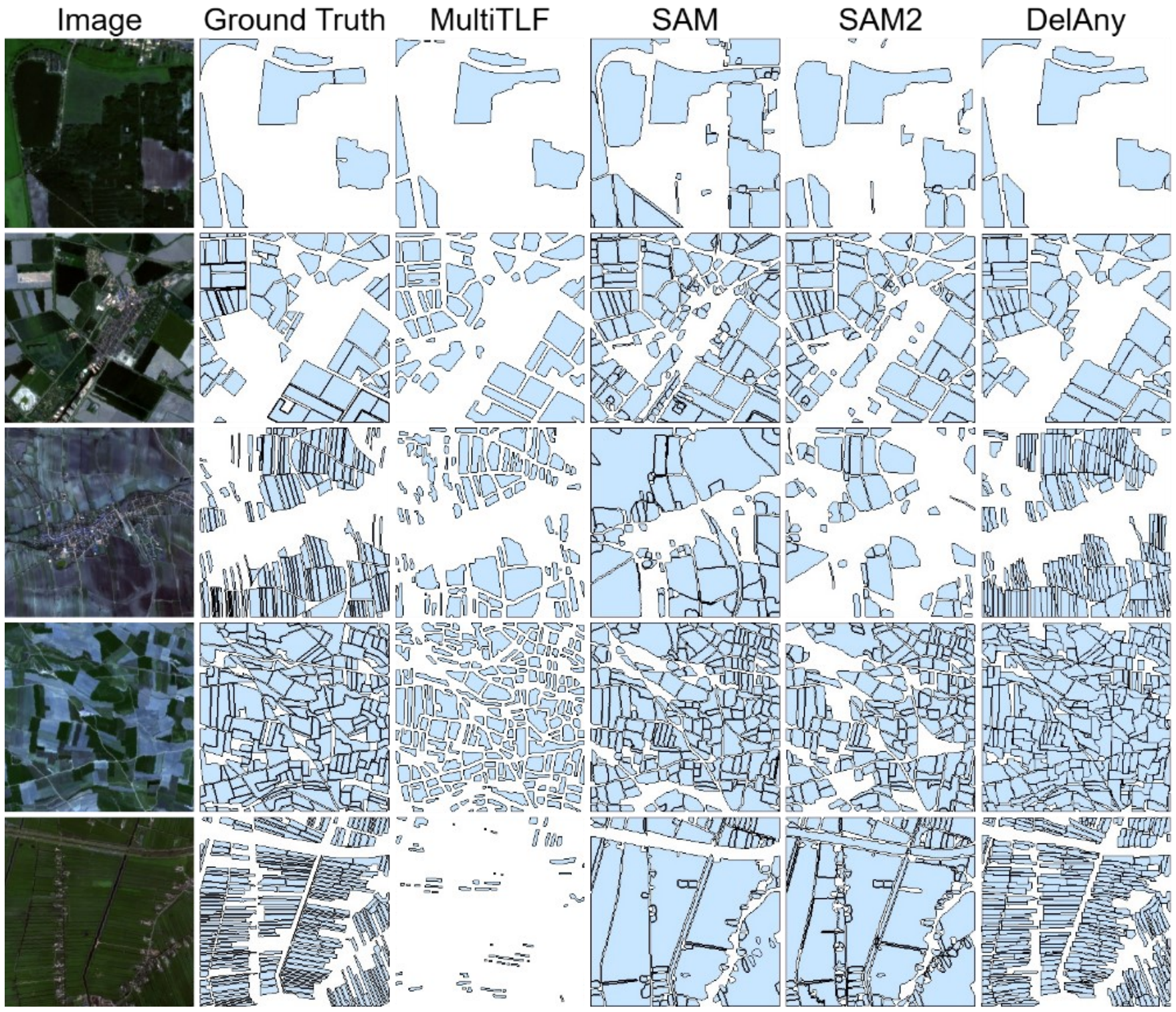

Figure 7. Qualitative results on the FBIS-22M test set. DelAny is compared to MultiTLF (Kerner et al., 2023), SAM (Kirillov et al., 2023), and SAM2 (Ravi et al., 2025). Different samples vary in size and density of the fields to illustrate model performance under diverse conditions.

### 3.3. Zero-Shot Cross-Region Feasibility

In the second experiment, we aimed to explore the possibility of applying the model to regions beyond the training distribution, so we applied the trained models to a geographically diverse set of regions absent from training in a strict zero-shot setting without fine-tuning: Brazil, Cambodia, New Zealand, Rwanda, the United States, Vietnam, and South Africa. These sites were selected to test model robustness across landscape structures, field sizes, and agricultural practices that differ from the European systems dominant in FBIS-22M. Because reference boundaries are unavailable for most of these regions, this experiment is qualitative by design.

The qualitative results (Figure 8) suggest that the model can be applied to regions beyond its training geography. Predictions remain coherent across smallholder farms, large industrial fields, and heterogeneous configurations. The model produces plausible results under conditions that typically challenge delineation models, such as irregular parcel shapes, inconsistent textures, and unfamiliar spatial layouts, capturing generalizable structural cues rather than region-specific patterns.

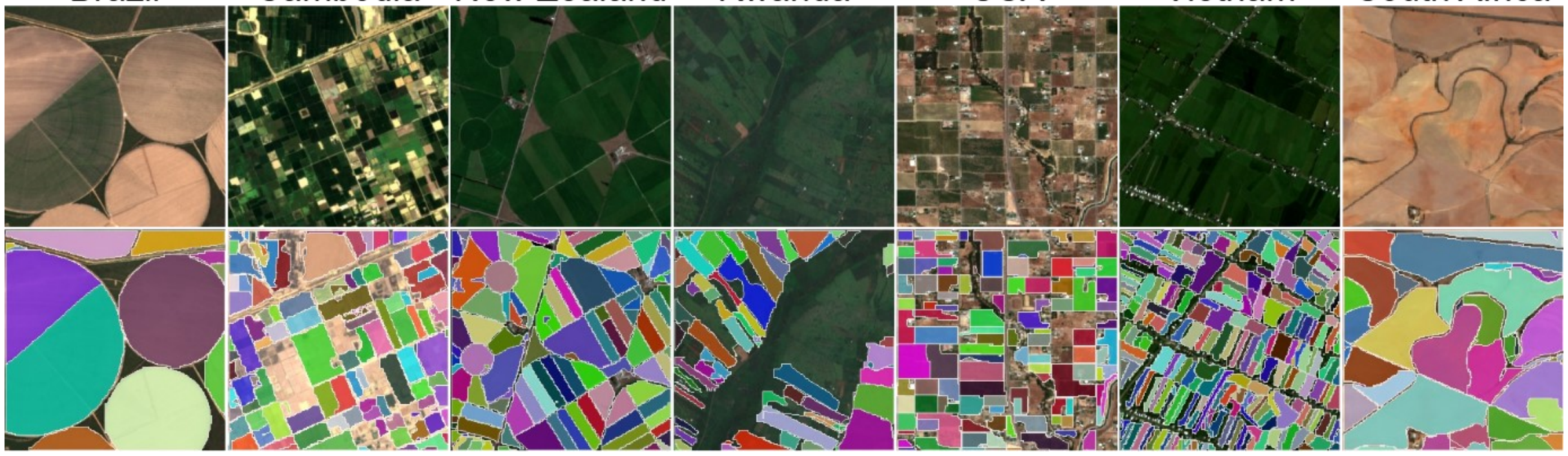

Figure 8. Qualitative results of zero-shot predictions. DelAny is applied to regions with different climates, terrains, and agricultural practices, highlighting its performance beyond the training data. Each color represents a different instance of the detected field.

### 3.4. Dataset Size and Diversity Impact

In the third experiment, to isolate the contribution of training data scale and diversity, DelAny was trained separately on AI4Boundaries (45K images), two FBIS-22M subsets (45K and 150K images), and the full FBIS-22M dataset (636K images), with performance compared using mAP@0.5 and mAP@[0.5:0.95] (Table 3).

Table 3. Impact of dataset size and diversity on model performance. Performance comparison of the DelAny model trained on the AI4Boundaries dataset and on different-sized subsets of FBIS-22M, highlighting the effect of dataset scale and diversity. Best results are in bold.

| Dataset | # Images | mAP@0.5 | mAP@0.5:0.95 |
|---|---|---|---|
| AI4Boundaries | 45K | 0.358 | 0.211 |
| FBIS-22M (subset) | 45K | 0.597 | 0.335 |
| FBIS-22M (subset) | 150K | 0.678 | 0.429 |
| **FBIS-22M** | 636K | **0.720** | **0.477** |

The results show consistent gains as the size of the FBIS-22M subset increases from 45K to 150K and finally to the full 636K images, with no indication of a performance plateau at the tested scales. Notably, when matched at the same image count (45K), the subset of FBIS-22M reveals a substantial performance advantage over AI4Boundaries. However, the two datasets also differ in resolution composition, annotation protocol, and geographic coverage, so matching image count alone does not fully control for these factors. The observed improvement is consistent with a contribution from greater dataset diversity, but differences in dataset composition, resolution, annotation protocol, and geographic coverage between AI4Boundaries and FBIS-22M may also play a role, and their relative contributions cannot be fully disentangled from this comparison alone.

### 3.5. Data Spatial Resolution Impact

In the fourth experiment, to evaluate the effect of input spatial resolution on delineation quality, we compared model performance across satellite sources of varying resolution: Sentinel-2 (native 10 m, plus cubic-interpolated 5 m and 2.5 m, and super-resolved 1 m), Planet (native 3 m), and Maxar (native 2 m and 30 cm, plus pan-sharpened variants), over 100 $km^2$ test regions in the Lvivska and Vinnytska oblasts of Ukraine. Lvivska, characterized by small, fragmented fields, supported qualitative assessment (Figure 9), while Vinnytska, where reference ground truth was available, enabled quantitative evaluation (Table 4).

Table 4. Impact of spatial resolution on field boundary delineation performance using Maxar, Planet, and Sentinel-2 imagery. Arrows indicate the desired direction of each metric. Best results are shown in bold.

| Image | ↑ IoU | ↑ small object recall | ↓ avg GOC | ↓ avg GUC | ↓ avg GTC |
|---|---|---|---|---|---|
| Maxar, 30 cm | **0.971** | **0.878** | **0.147** | 0.898 | 0.673 |
| Maxar, 1 m | 0.968 | 0.858 | 0.184 | 0.884 | 0.673 |
| Maxar, 3 m | 0.953 | 0.753 | 0.276 | 0.908 | 0.714 |
| Maxar, 5 m | 0.945 | 0.713 | 0.303 | 0.923 | 0.731 |
| Planet, 3 m | 0.936 | 0.693 | 0.299 | **0.825** | **0.659** |
| S2, 1 m | 0.931 | 0.812 | 0.300 | 0.842 | 0.668 |
| S2, 10 m | 0.884 | 0.102 | 0.775 | 0.932 | 0.884 |

Across both regions, delineation quality improves with finer spatial resolution, with the clearest gains concentrated in the detection and separation of small fields. Native Maxar imagery yields the strongest overall performance, and accuracy degrades only marginally down to 1 m. Degradation becomes more pronounced at 3 m and 5 m, where boundary-error metrics increase alongside a decrease in IoU, while native Sentinel-2 at 10 m shows the sharpest decline, particularly in small-object recall, reflecting a loss of field-level separability for small, fragmented parcels.

The qualitative comparison further illustrates these trends. Although the overall field structure remains consistent across Sentinel-2 resolutions (Figure 9c–e), cubic interpolation improves the detection of small private fields. Interpolated Sentinel-2 at 2.5 m produces delineation results approaching those of native 3 m Planet imagery (Figure 9e, f), with Planet's remaining advantage attributable primarily to its higher native image sharpness rather than nominal spatial resolution alone. Native Maxar imagery provides the most detailed delineation (Figure 9g, h), although at the highest resolutions it exhibits a tendency to merge adjacent small fields into larger polygons.

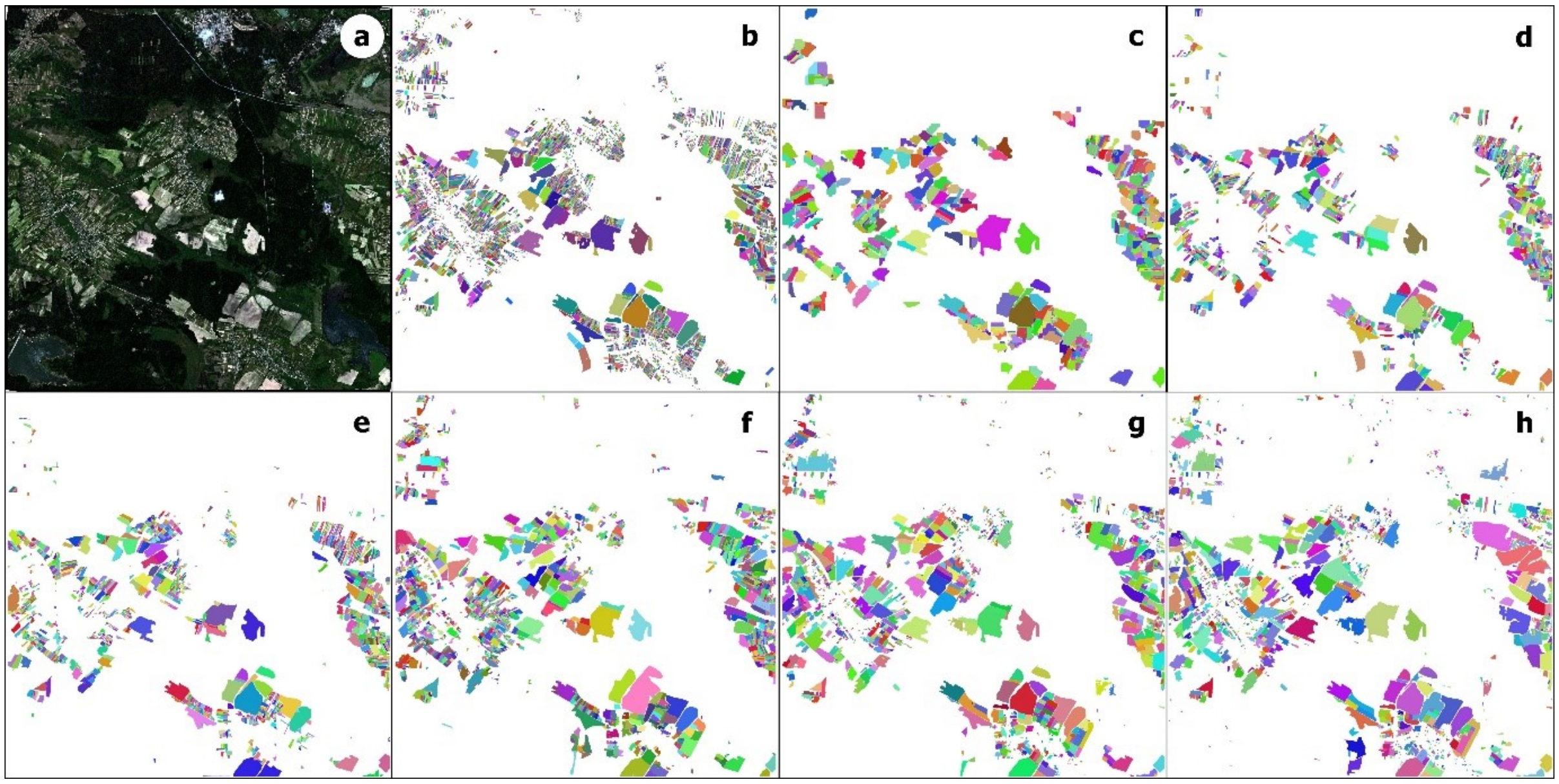

Figure 9. Field delineation over a 100 km² area in Lvivska oblast (patch centre: 23.3877, 49.8848) from imagery of different spatial resolution: (a) Maxar image; (b) manually delineated fields; (c) Sentinel-2, 10 m; (d) Sentinel-2 interpolated to 5 m; (e) Sentinel-2 interpolated to 2.5 m; (f) PlanetScope, 3 m; (g) Maxar, 2 m; (h) Maxar pan-sharpened to 0.5 m. Each colour represents a different detected field instance.

Manual inspection showed that the minimum reliably delineated field size decreases with increasing spatial resolution, from approximately 0.5 ha for Sentinel-2 to 0.05 ha for pan-sharpened Maxar. This trend is mirrored in the log-scaled area distributions (Figure 10), capturing the high retrieval rate of micro-parcels at sub-meter resolutions, balanced by a localized over-segmentation tendency in Maxar data within the 10–70 ha range. Cubic interpolation of freely available Sentinel-2 imagery substantially narrows the performance gap to native commercial sub-5 m imagery while requiring no additional data acquisition, providing a practical, low-cost pathway toward accurate field delineation in regions where access to commercial imagery is limited.

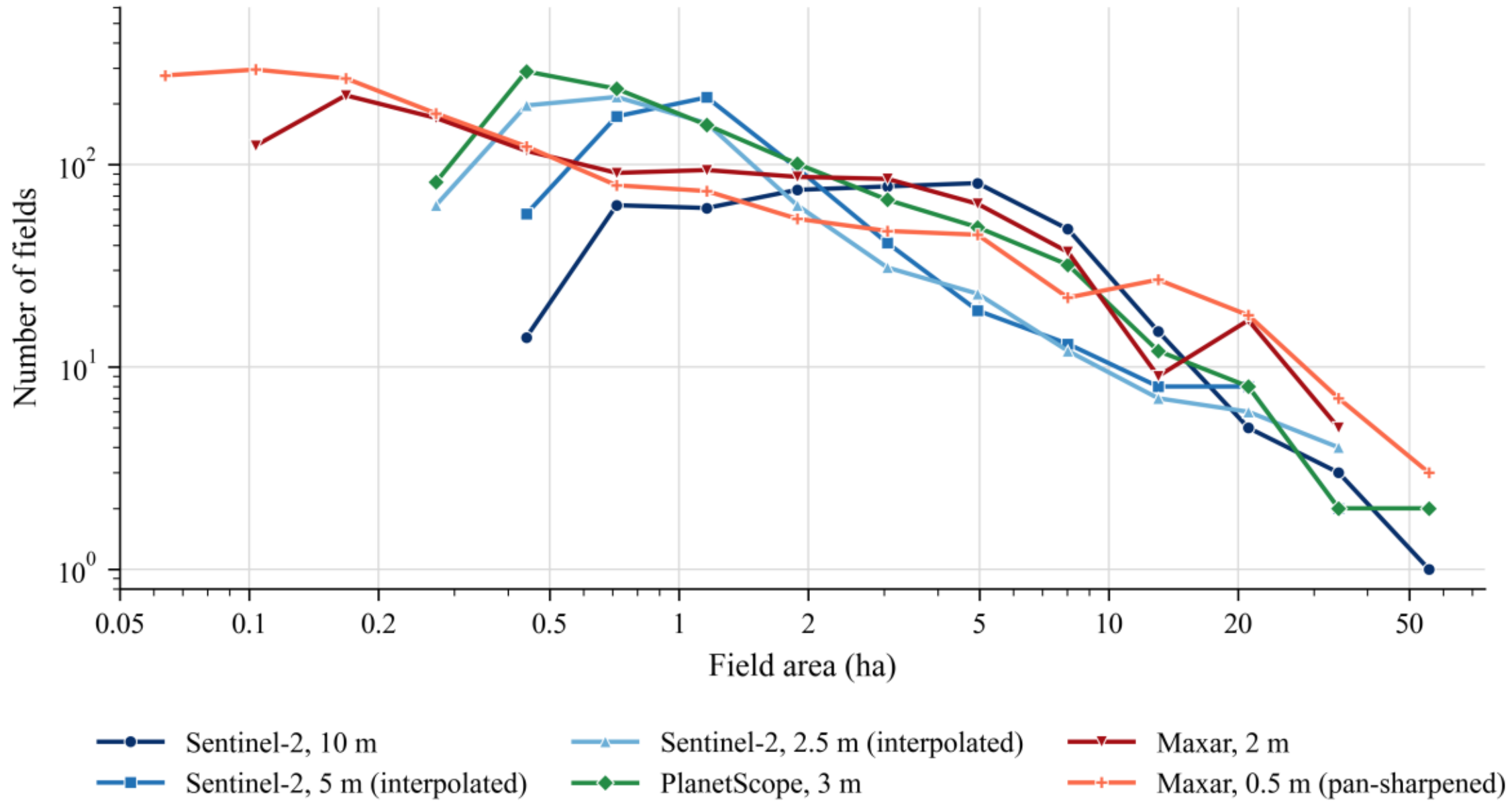

Figure 10. Number of detected fields per field-area class (logarithmic bins) in the Lvivska oblast test area for imagery from different sensors and spatial resolutions. Fields below the minimum size reliably delineated at each resolution are not shown (0.5 ha for Sentinel-2 at 10 and 5 m, 0.3 ha for Sentinel-2 at 2.5 m and PlanetScope, 0.1 ha for Maxar at 2 m).

### 3.6. Within-Growing-Season Acquisition-Date and Preprocessing Sensitivity

The fifth experiment investigated within-growing-season acquisition-date sensitivity using PlanetScope imagery (3 m resolution) over a 100 km² block in Vinnytska oblast (Ukraine), comparing four acquisition dates spanning two growing seasons and multiple phenological stages (6 May 2022, 4 July 2022, 21 May 2023, 28 July 2023; Table 5, Figure 11). Within this spring–summer window, performance proves largely insensitive to acquisition date: IoU remains between 0.931 and 0.936 regardless of season or year, and boundary-disagreement metrics vary marginally across dates. Small object recall shows somewhat greater spread (0.657 to 0.742), reflecting phenological effects on canopy structure and inter-field spectral contrast, but no single acquisition date stands out as systematically better or worse, confirming robustness to seasonal timing.

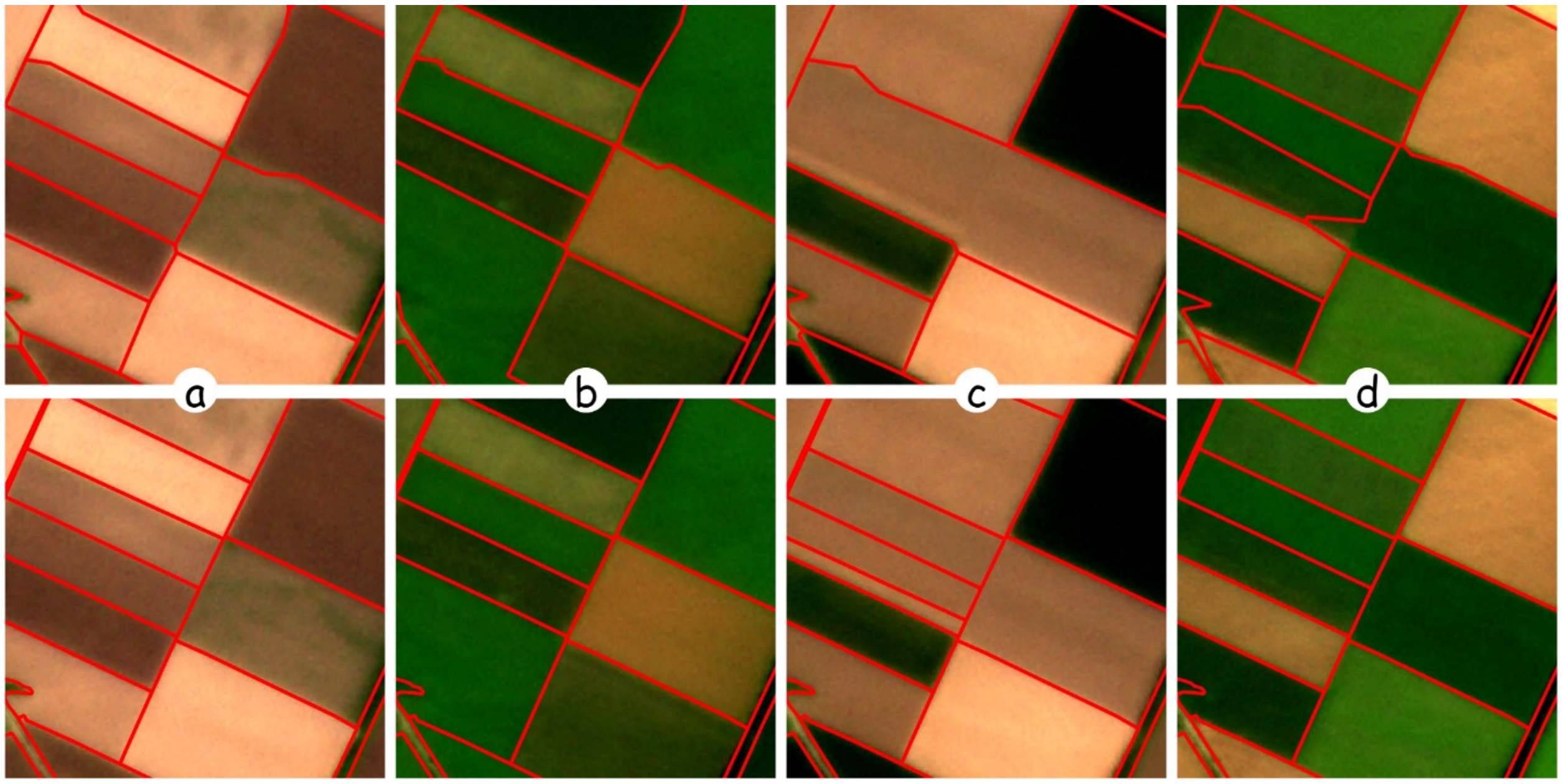

Figure 11. Predicted field boundaries (top row) and reference boundaries (bottom row) overlaid on PlanetScope imagery for the same area across four acquisition dates: (a) 6 May 2022, (b) 4 July 2022, (c) 21 May 2023, and (d) 28 July 2023.

Table 5. Impact of within-growing-season acquisition date (May–July, 2022–2023) on field boundary delineation performance using PlanetScope 3 m imagery. Arrows indicate the desired direction of each metric.

| Image | ↑ IoU | ↑ small object recall | ↓ avg GOC | ↓ avg GUC | ↓ avg GTC |
|---|---|---|---|---|---|
| 21 May 2023 | 0.936 | 0.693 | 0.299 | 0.825 | 0.659 |
| 4 July 2022 | 0.932 | 0.657 | 0.338 | 0.827 | 0.671 |
| 6 May 2022 | 0.933 | 0.701 | 0.315 | 0.837 | 0.672 |
| 28 July 2023 | 0.931 | 0.742 | 0.297 | 0.846 | 0.673 |

The sixth experiment examined whether alternative spectral or temporal preprocessing of Sentinel-2 imagery could bridge the resolution-driven performance gap. Using Sentinel-2 data from the same area, we compared the standard growing-season RGB composite with single-band multi-date composites (blue, green, red, and near-infrared, each built from three within-season acquisitions), a cross-channel temporal RGB composite, and a harmonic time-series composite based on the work of Papić et al. (2025) (Table 6). None of these alternatives meaningfully outperformed the baseline. The Red and Green single-band composites edge out the standard composite marginally, while the temporal RGB and harmonic composites perform worse across all metrics. This indicates that encoding seasonal variation across channels does not substitute for genuine spatial resolution. Classical contrast-enhancement techniques improved visual sharpness but produced no measurable gain in quantitative delineation quality.

Table 6. Impact of preprocessing on field boundary delineation performance. Each S2 variant represents a composite of 3 dates for the corresponding spectral channel. Arrows indicate the desired direction of each metric. Best results are shown in bold.

| Image | ↑ IoU | ↑ small object recall | ↓ avg GOC | ↓ avg GUC | ↓ avg GTC |
|---|---|---|---|---|---|
| S2 | 0.884 | 0.102 | 0.775 | 0.932 | 0.884 |
| Blue, 3 dates | 0.879 | 0.056 | 0.829 | 0.918 | 0.892 |
| Green, 3 dates | 0.891 | 0.117 | 0.756 | 0.913 | 0.866 |
| Red, 3 dates | **0.896** | **0.136** | **0.733** | **0.908** | **0.856** |
| NIR, 3 dates | 0.868 | 0.026 | 0.867 | 0.921 | 0.908 |
| RGB, 3 dates | 0.869 | 0.047 | 0.846 | 0.922 | 0.900 |
| Harmonic composite | 0.842 | 0.008 | 0.905 | 0.929 | 0.924 |

### 3.7. Comparison with Other Global-Scale Products

#### 3.7.1. Quantitative Results

To compare the quality of our model's delineation with existing products, including NASA Harvest, Sinergise Solutions, and Fields of the World (FTW), we evaluated all methods using an independent Land Parcel Identification System (LPIS) test site modified to enforce strict single-field object verification. Each agricultural field was delineated as an individual object, and edge-intersecting fields were retained. Ground truth was derived from manual delineation on 30 cm Maxar imagery, with a separate Planet-resolution reference constructed to enable comparison at that resolution (Table 7).

Across this evaluation, DelAnyFlow outperformed all operational products on every metric, with its Maxar-based configuration achieving the strongest overall result (IoU = 0.971). DelAnyFlow substantially reduces boundary-disagreement error and shows higher sensitivity to small, fragmented parcels. FTW is the closest competitor overall, while NASA Harvest and Sinergise Solutions lag further behind, particularly on small-object detection. Direct runtime comparison against these products was not possible, as their processing infrastructure is not publicly documented.

Table 7. Quantitative comparison of field boundary delineation products on the LPIS test site. Arrows indicate the desired direction of each metric. Best results are shown in bold.

| Product | Source | ↑ IoU | ↑ small object recall | ↓ avg GOC | ↓ avg GUC | ↓ avg GTC |
|---|---|---|---|---|---|---|
| NASA Harvest | Planet | 0.850 | 0.000 | 0.926 | 0.953 | 0.945 |
| Sinergise Solutions | Sentinel-2 | 0.834 | 0.080 | 0.861 | 0.915 | 0.902 |
| FTW | Sentinel-2 | 0.926 | 0.405 | 0.520 | 0.973 | 0.831 |
| **DelAnyFlow** | **Planet** | 0.936 | 0.693 | 0.299 | **0.825** | **0.659** |
| **DelAnyFlow** | **Maxar** | **0.971** | **0.878** | **0.147** | 0.898 | 0.673 |

#### 3.7.2. Operational Deployment

Beyond controlled benchmarking, we assessed operational feasibility at national scale by generating a full 2024 field boundary layer for Ukraine using resampled Sentinel-2 imagery at 5 m and 2.5 m. Ukraine, with an area of 603,000 km², features considerable agricultural diversity, ranging from large, contiguous plots to fragmented micro-fields.

The full map was generated in 5.4 hours on a consumer-grade desktop (AMD Ryzen 9 9900X 12-core CPU, NVIDIA GeForce RTX 5070 Ti 16 GB GPU, and 64 GB RAM), excluding data acquisition, demonstrating the computational feasibility and scalability of the approach using accessible hardware. There is no comparable runtime data for existing global-scale products, so direct operational speed comparisons could not be performed. Figure 12 shows the input Sentinel-2 composite (imagery from the growing season from May 15 to August 15, 2024) and the resulting delineation output.

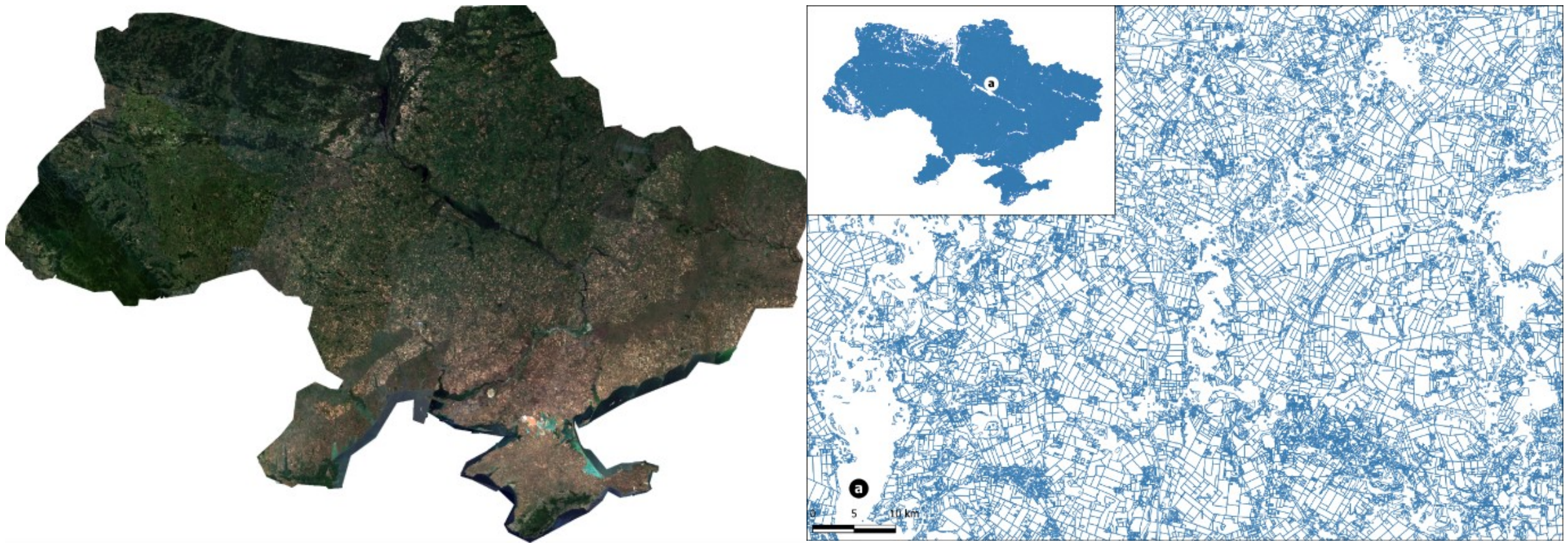

(a) (b)

Figure 12. (a) Input Sentinel-2 composite over Ukraine; (b) Output vector fields.

We compared our results with those from Sinergise Solutions (2024) and NASA Harvest (2023). Figure 13 shows the distribution of fields larger than 0.25 hectares, which is the practical lower limit that can be reliably delineated from Sentinel-2 imagery (~25 pixels per field). Our method detected 3.75 million fields using interpolated 5 m data and 5.15 million fields with 2.5 m data, compared to 2.66 million for Sinergise Solutions and 1.69 million for NASA Harvest (2023 data). A higher field count is not by itself evidence of superior quality, since over-segmentation can also inflate counts. Therefore, these figures are descriptive and should be read alongside the quantitative results in Table 7.

Figure 13 shows that our method produces more fields in the 1,000–10,000 ha range than the other products, and detects substantially more fields in the 1–10 ha range. The 2.5 m configuration also improves detection of small fields (0.25–1 ha), but its rate of large-field over-detection and longer processing time make the 5 m version more practical for operational use.

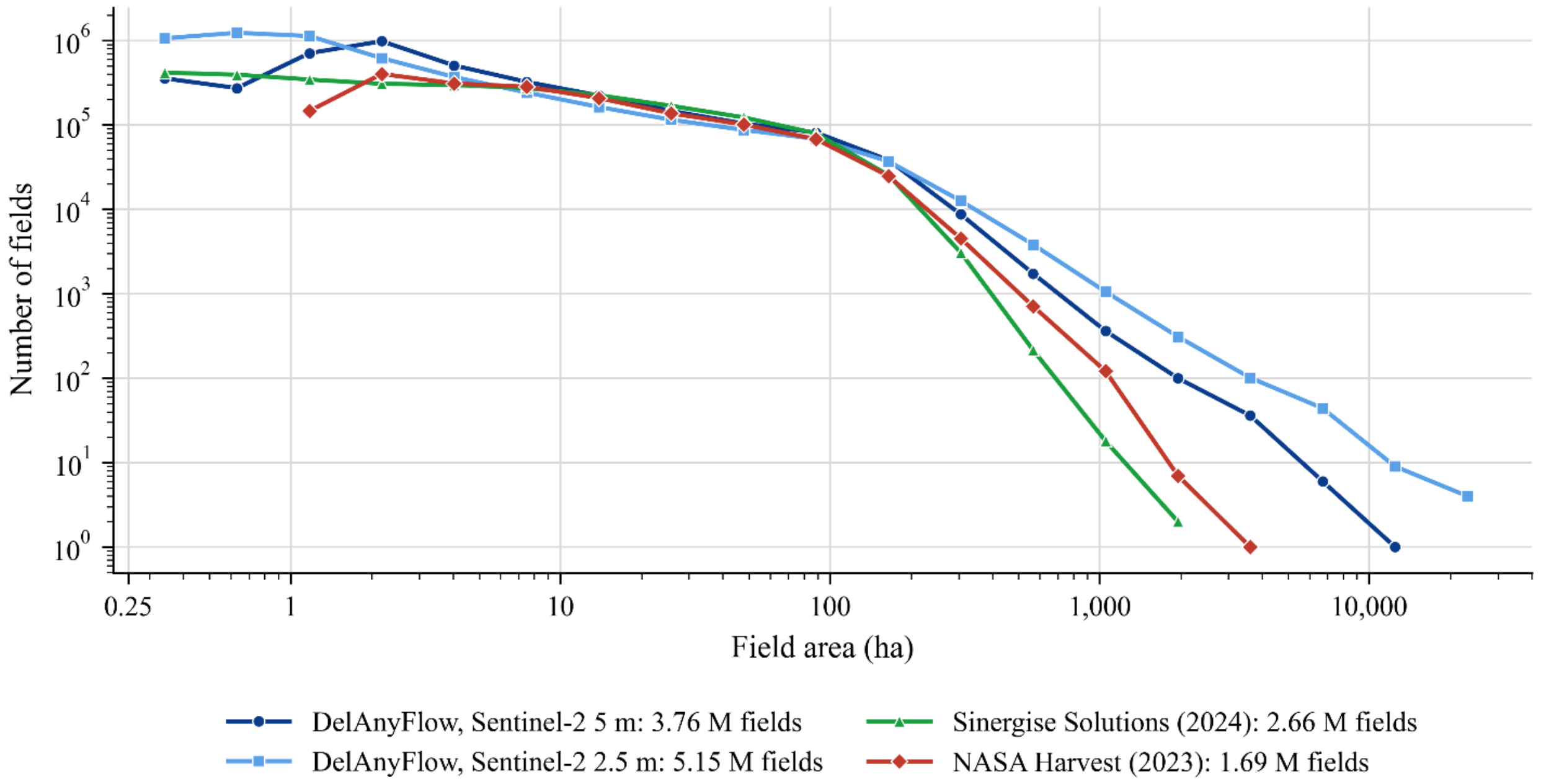

Figure 13. Number of fields per field-area class (logarithmic bins, fields of at least 0.25 ha) for the DelAnyFlow maps of Ukraine and the Sinergise Solutions and NASA Harvest products.

### 3.7.3. Geometric and Topographic Analysis

Figure 14 shows a comparison of delineations across four regions with an area of approximately 7 km². DelAnyFlow detects small private fields more effectively than competing methods: Sinergise Solutions captures only a portion of these parcels, while NASA Harvest fails to detect them altogether. For medium-sized fields, DelAnyFlow and Sinergise Solutions perform comparably, and both outperform NASA Harvest. FTW covers a similar agricultural extent and retains many small fields, but frequently merges adjacent parcels into single polygons and occasionally misclassifies non-agricultural areas (urban zones, bare land, grassland) as fields.

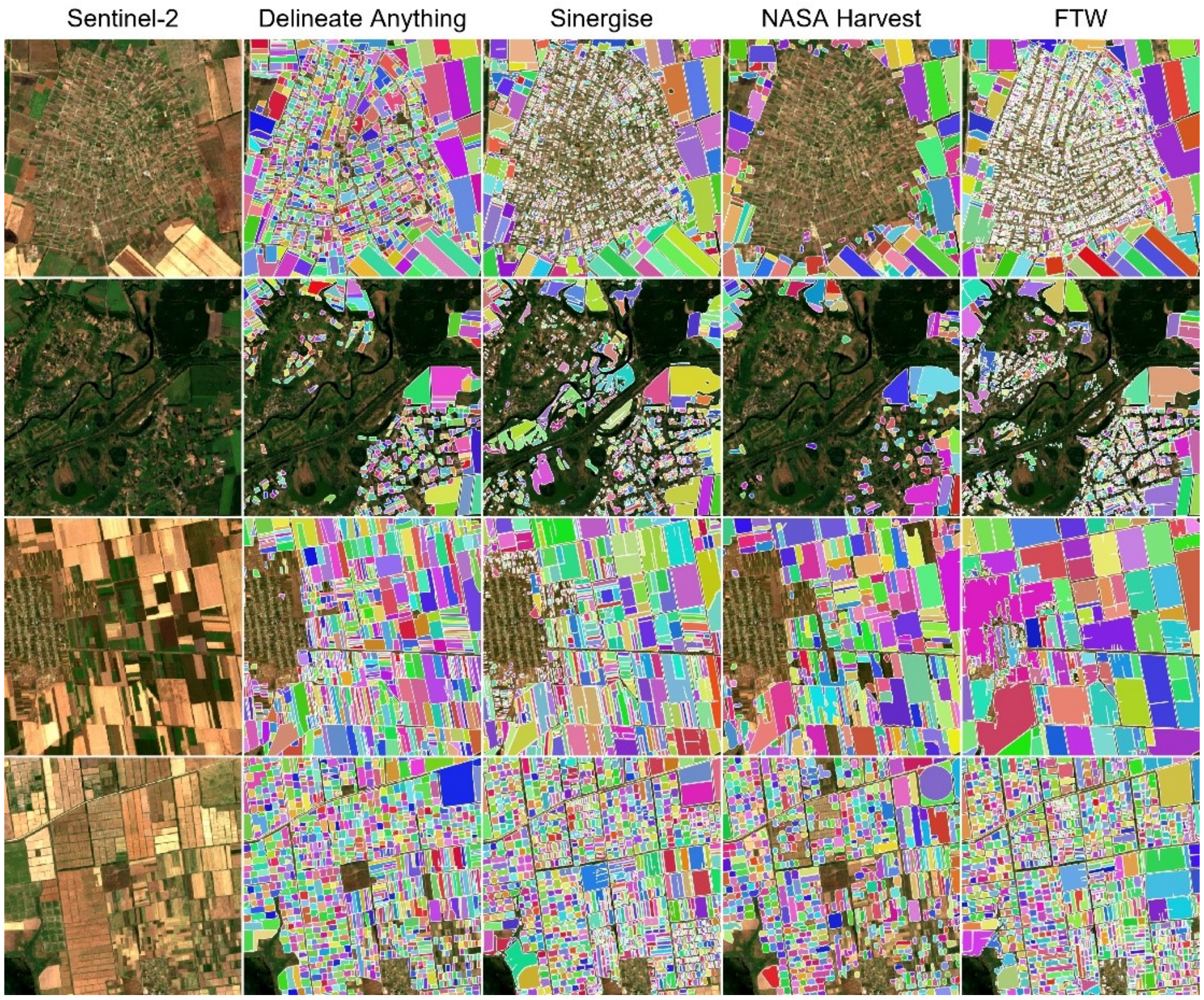

Figure 14. Comparison of field delineations from different products across four 7 km² regions. Each color represents a different instance of the detected field.

Unlike Sinergise Solutions, which applies a buffering step that leaves spatial gaps between plots, DelAnyFlow's field boundaries more accurately reflect true physical extent, making them better suited to downstream analyses requiring precise area estimates. Figures 15 and 16 provide close-up views of two sites using Maxar imagery. At the first site (Figure 15), DelAnyFlow accurately delineates individual plots and preserves the internal structure. In contrast, NASA Harvest merges most parcels into a few large polygons, ignoring nearly all internal boundaries. Sinergise Solutions detects the main blocks but under-segments the internal parcels and leaves gaps due to buffering, while FTW aggregates the entire area into a single polygon. At the second site (Figure 16), DelAnyFlow identifies the narrow plots on the left side of the scene. NASA Harvest merges these plots and fails to detect them. Sinergise Solutions separates the major fields, misses the smallest parcels, and leaves incomplete polygons, while FTW groups most of the area into large blocks, losing the internal field structure visible in the imagery.

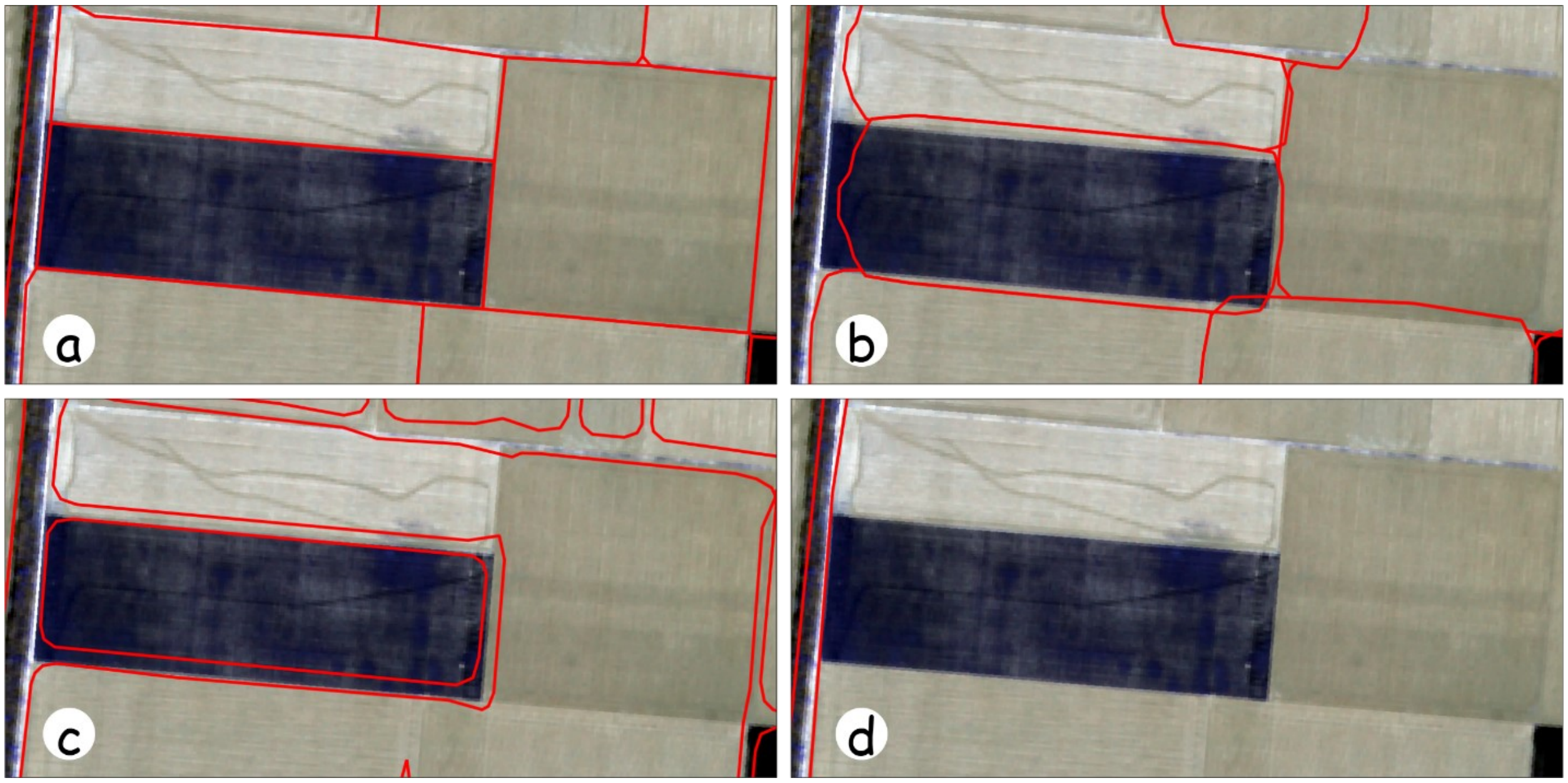

Figure 15. Close-up comparison of field boundary delineation quality (site 1). Maxar imagery with overlaid field boundaries for four products: (a) DelAnyFlow, (b) NASA Harvest, (c) Sinergise Solutions, and (d) Fields of the World.

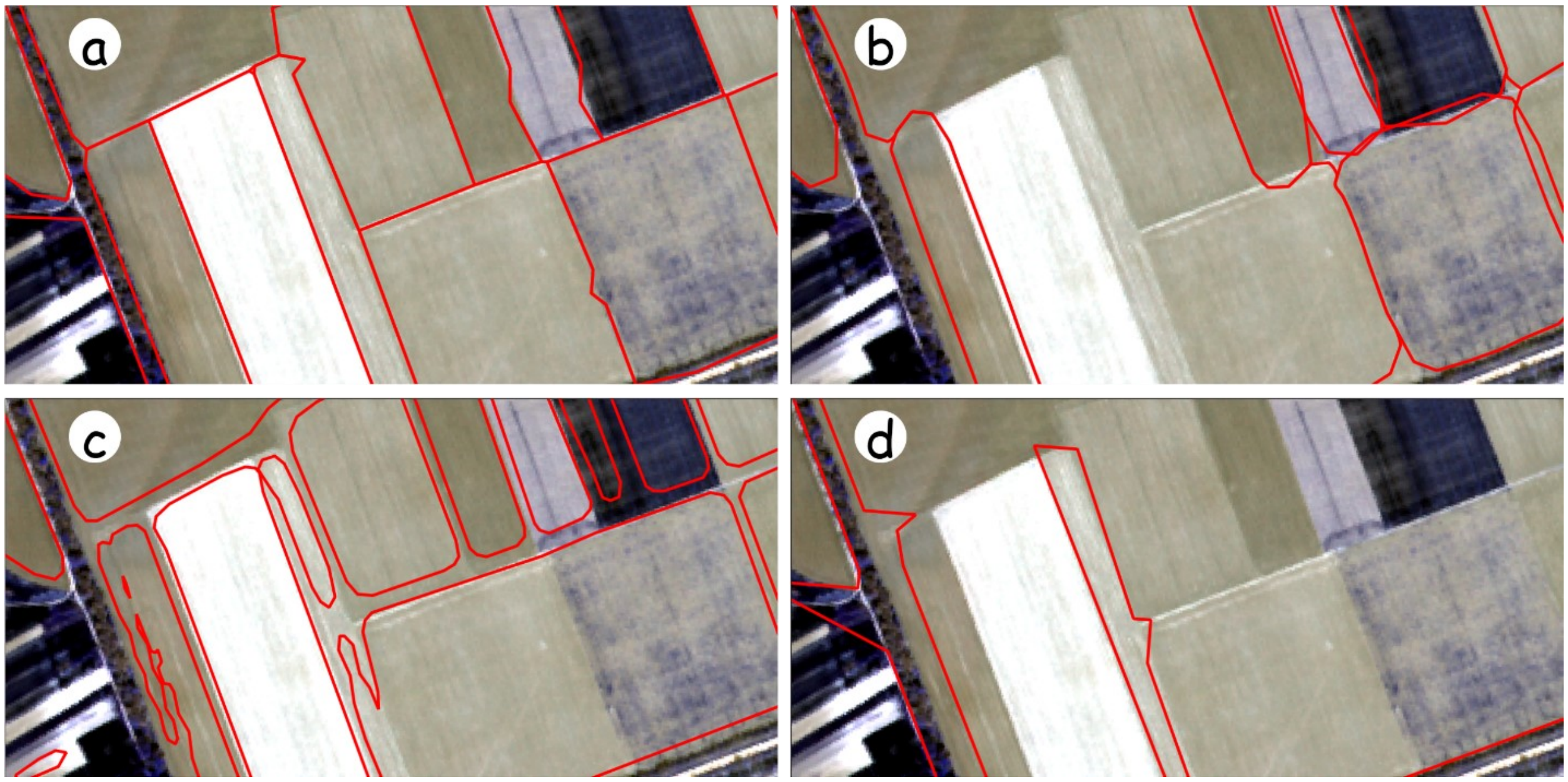

Figure 16. Close-up comparison of field boundary delineation quality (site 2). Maxar imagery with overlaid field boundaries for four products: (a) DelAnyFlow, (b) NASA Harvest, (c) Sinergise Solutions, and (d) Fields of the World.

In addition to detecting more individual fields, DelAnyFlow produces clean boundary geometry. The simplified field contours preserve sharp edges and characteristic shapes, matching how human delineators digitize them, unlike the overly smoothed or irregular contours observed in other operational products. A broader side-by-side visual comparison of delineation results across multiple regions is available in Supplementary Video 1 (Figure 17).

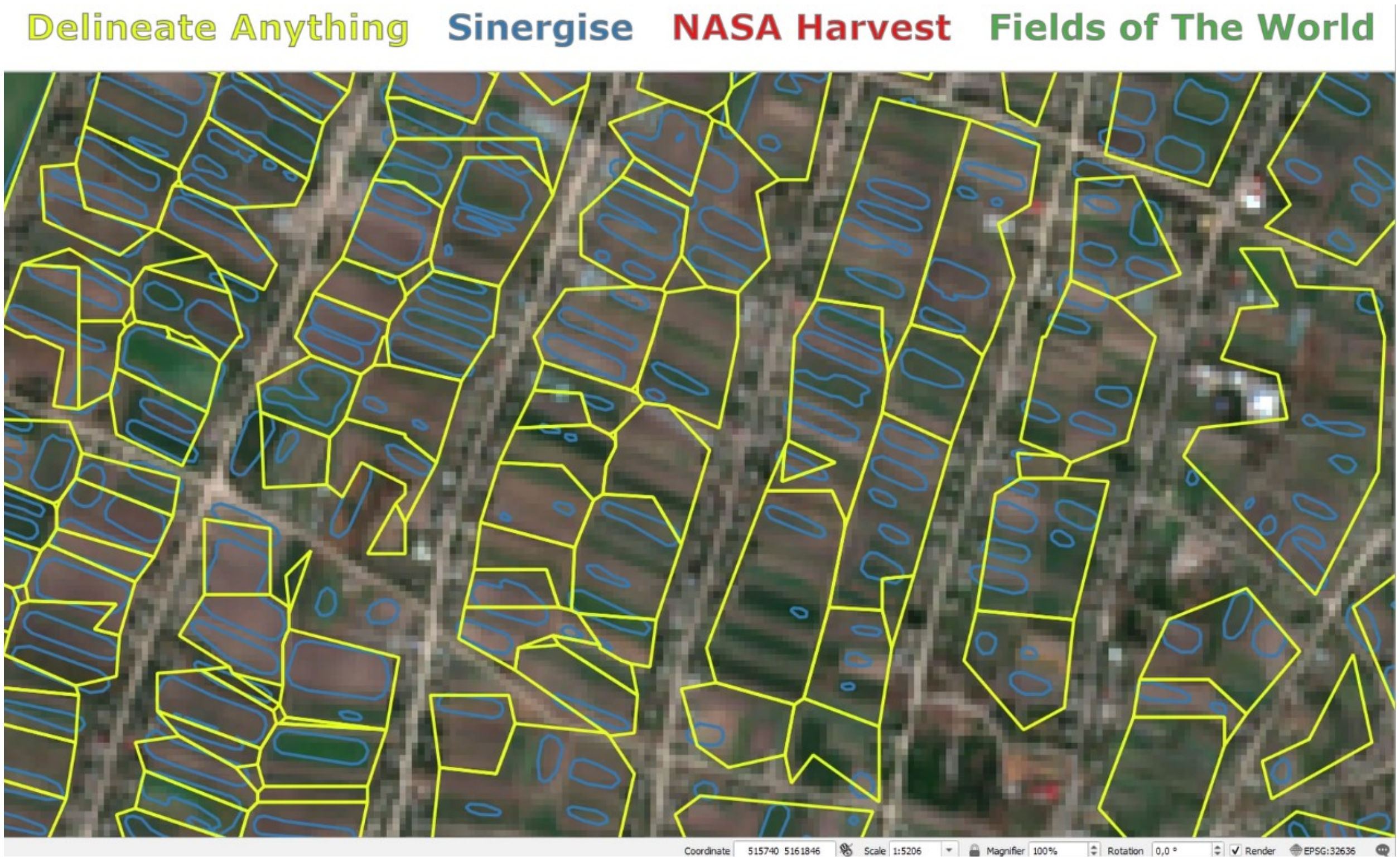

Figure 17. Supplementary Video 1: Comparative delineation results across several regions using DelAnyFlow, Sinergise Solutions, NASA Harvest and Fields of the World.

# 4. Discussion

## 4.1. Methodological Advances and Remote Sensing Benchmarking

The results of this study demonstrate that reformulating field boundary delineation as an instance segmentation problem substantially improves both geometric accuracy and computational scalability compared to conventional semantic segmentation approaches (Diakogiannis et al., 2020; Waldner et al., 2021). By leveraging the massive geographic and spatial diversity of the FBIS-22M dataset, the DelAny model achieves strong performance on regions outside its training distribution (Section 3.3), though this evidence remains qualitative. Delivering inference speeds up to 400× faster than computer vision foundation models like SAM2 (Ravi et al., 2025), the proposed architecture avoids the computational overhead and extensive post-processing typically required by multi-stage semantic segmentation networks (Waldner et al., 2021).

To clarify the spatial assumptions of this framework, we do not claim strict scale invariance. Ground sampling distance fundamentally alters how features are represented. Medium-resolution imagery (10 m) introduces sub-pixel mixing near narrow boundaries, which prevents reliable parcel discrimination (Wang et al., 2022). This is consistent with the longstanding evidence that field-parcel detectability depends on resolution (Graesser and Ramankutty, 2017). Furthermore, sub-meter commercial imagery introduces additional high-frequency textures, such as machinery rows and individual canopy variation. In this context, our resolution-agnostic paradigm defines the capacity of a single operational pipeline to process multi-sensor inputs from 0.25 to 10 m without structural or parametric modifications. Benchmarking against established operational Earth Observation (EO) products from Sinergise Solutions, NASA Harvest, and Fields of The World (Kerner et al., 2024a) indicates that DelAnyFlow extracts a significantly higher proportion of fields in the critical range from 1 to 10 ha. Quantitative evaluations across IoU, small object recall, and boundary discrepancy metrics (avg GOC, GUC, GTC) confirm that our instance-level formulation effectively minimizes parcel merging while preserving sharp, realistic contours that closely match human-digitized limits.

### 4.2. Sources of Uncertainty, Phenological Factors, and Methodological Boundaries

Despite promising qualitative evidence of cross-region applicability, several methodological limitations and sensor-specific constraints must be formalized. First, the FBIS-22M training data is predominantly compiled from European landscapes, which introduces a geographical selection bias toward temperate, high-input monocultures. Direct transferability to highly fragmented smallholder farms in sub-Saharan Africa or Southeast Asia requires systematic evaluation against localized datasets (Persello et al., 2023). Second, instance segmentation introduces distinct spatial challenges in continuous geographical spaces compared to pixel-wise semantic classification. Matching and merging discrete field instances across adjacent image tiles can generate boundary duplicates or artificial edge truncations, necessitating our dedicated topology-consistent post-processing pipeline.

A primary radiometric trade-off in this study is the exclusive reliance on RGB-like bands. This selection was strictly mandated to ensure maximum cross-sensor compatibility, though it omits vital multispectral data such as the Near-Infrared (NIR) and RedEdge bands that are essential for separating phenologically identical adjacent crops (Sitokonstantinou et al., 2018). Consequently, when spectral boundaries are weak or cloud-induced data gaps occur, DelAnyFlow can exhibit over-aggregation, occasionally producing irregular polygons exceeding 1000 ha. This underscores that field edge visibility is fundamentally governed by seasonal crop development and local management practices (Sadeh et al., 2025). Our acquisition-date experiment (Section 3.6) was similarly limited to the spring–summer growing season. A systematic evaluation across autumn and winter conditions, where bare soil, residue, and snow cover can substantially alter boundary contrast, remains untested and is left to future work. Furthermore, the current workflow lacks an internal uncertainty quantification mechanism, meaning that spatial confidence estimation and error propagation into downstream agricultural statistics remain unaddressed.

### 4.3. Framework Extensibility and Policy Implications

DelAnyFlow's core model operates on three-band RGB imagery. Since the training data spans multiple satellites and spatial resolutions, the model can be applied to any optical source that provides RGB bands. As demonstrated in Section 3.6, the model can also be applied to inputs rendered as RGB composites from time series. The model can also be applied to non-optical Earth observation (EO) data, such as synthetic aperture radar (SAR) (Diakogiannis et al., 2024) or hyperspectral data, provided the user supplies an RGB-like rendering of the data. Supplementary Section S4 shows examples of delineation without retraining from a hue-shifted Sentinel-1 SAR composite, a harmonic NDVI composite, and a projection of TESSERA-learned embeddings. Supplementary Section S5 quantifies the zero-shot transfer of DelAny on an independent benchmark of 100 countries, most of which are absent from FBIS-22M, and the released model has already been applied without fine-tuning to LPIS creation in Albania and Ukraine (Luketić et al., 2026).

While lacking the sub-meter tolerances and manual control of EU LPIS systems (Scown et al., 2020), the workflow provides an operational baseline for nations without digital cadastral registries. Plot-level aggregation mitigates the mixed-pixel noise that typically degrades classification in fragmented landscapes (Peña-Barragán et al., 2011). Operational vector outputs are well suited to inform downstream applications such as conflict-related agricultural damage assessments (Kussul et al., 2023), environmental governance (Hall et al., 2021), and farm-size distribution mapping for food security policies. Future research will prioritize expanding training dataset coverage, embedding conformal prediction for robust uncertainty estimation, and scaling via cloud-native, API-driven implementations.

## 5. Conclusion

This study presented DelAnyFlow, a resolution-agnostic instance segmentation framework for generating topologically consistent agricultural boundaries from multi-resolution satellite imagery. Cross-tile field unification and topology-consistent simplification substantially reduce tile-boundary fragmentation while preserving field geometry. DelAnyFlow thus extends tile-level predictions into a scalable country-scale geospatial mapping framework.

Because FBIS-22M training data are predominantly European, transferability to underrepresented smallholder landscapes such as sub-Saharan Africa and Southeast Asia should be confirmed against localized reference data before operational deployment in those regions. The resulting layers can inform downstream applications such as plot-level crop classification, yield estimation, and food security assessments across both smallholder and industrial farming systems, though these applications fall outside the scope of the delineation experiments presented here.

To ensure reproducibility and facilitate integration into GEOGLAM, FAO, and national statistical frameworks, all trained models, source code, and vector layers are publicly available. This scalable, open-source workflow provides a verifiable baseline for next-generation agricultural intelligence and spatial analysis, with the Ukraine case study demonstrating feasibility at national scale.

## Acknowledgements

This work was partially supported by the European Space Agency (ESA), the European Commission through the joint World Bank/EU project 'Supporting Transparent Land Governance in Ukraine' [grant numbers ENI/2017/387-093 and ENI/2020/418-654], the Ministry of Education and Science of Ukraine through the project "Next-Generation Copernicus Services: An Intelligent Digital Twin for Forest Cover Analysis in Ukraine" [State Registration No. 0126U001302], the Swiss National Science Foundation (SNSF) through the project "DT4LC – Developing Scalable Digital Twin Models for Land Cover Change Detection Using Machine Learning" [grant number IZURZ2_224912] (development and validation of training data), and NASA funded project "Detecting and Mapping War-Induced Damage to Agricultural Fields in Ukraine Using Multi-Modal Remote Sensing Data" [grant number 80NSSC24K0354].

We thank Sinergise Solutions and NASA Harvest for providing access to their field boundary datasets for Ukraine used in the comparative evaluation, and Dr. Bohdan Yailymov (Space Research Institute NASU-SSAU) for his contribution to the automated formation of the FBIS-22M dataset.

## Data availability

The DelAny model code, weights, and inference pipeline are hosted at https://github.com/Lavreniuk/Delineate-Anything, and the FBIS-22M dataset subset is accessible at https://huggingface.co/datasets/MykolaL/FBIS-22M. The 2024 national-scale field boundaries for Ukraine are visualized at https://explorer.delineate-anything.apex.esa.int, where minor local spatial discrepancies with the high-resolution basemap may occur due to differing acquisition dates and sensor resolutions.

## Supplementary material

The following tables report the default DelAnyFlow parameters used in all experiments unless otherwise stated. These values were determined empirically during pipeline development and were subsequently fixed for the experiments reported in this study. For more detail information about parameters please check GitHub: https://lavreniuk.github.io/Delineate-Anything/.

Table S1. DelAnyFlow execution parameters

| Parameter | Default value | Description |
|---|---|---|
| tile_step | 0.5 (256px) | Controls the overlap between adjacent tiles. A value of 1 corresponds to no overlap, whereas 0.5 corresponds to 50% overlap. A value of 0.5 provides a good balance between delineation completeness and computational efficiency. A value of 1 is not recommended because it may result in visible seams between tiles. Values below 0.5 substantially increase execution time without providing a notable improvement in output quality, whereas values above 0.5 may reduce field recall. |
| use_half | false | Controls the use of FP16 precision during model inference, when supported by the hardware. Enabling FP16 generally reduces inference time with a small potential decrease in delineation accuracy. |
| minimal_confidence | 0.005 | Minimum confidence score required for retaining model predictions. This is the most important model-dependent parameter and may require adjustment for different target regions. Lower values generally increase delineation recall but also increase computational time and memory consumption. Lower confidence thresholds may additionally improve the detection of smaller fields and large partial fields located near tile boundaries. |
| super_resolution | null | Specifies the upscaling factor applied to the input imagery. When set to null, the factor is selected automatically: 1× for imagery with spatial resolution below 5 m and 2× for imagery with spatial resolution above 5 m. Other supported values include 1×, 2×, 4×, |
| | | and higher factors. The image is rescaled using cubic interpolation. |

Table S2. DelAnyFlow postprocessing parameters

| Parameter | Default value | Description |
|---|---|---|
| pixel_area_threshold | 512 px | Minimum area of a detected instance required for further processing. Lower values may increase the recall of small fields but may also reduce delineation quality because smaller instances are generally less reliable. For Sentinel-2 imagery processed with 2× super-resolution, a value of 512 pixels corresponds to approximately 1.28 ha. |
| remaining_area_threshold | 0.8 | Minimum fraction of the original instance that must remain after morphological processing for the instance to be retained. This parameter helps exclude instances composed of multiple disconnected regions or instances with substantial overlap with land-cover classes specified for clipping or filtering. |
| compose_merge_iou | 0.8 | Minimum IoU between two instances required for merging during tile composition. Higher values reduce the number of merging operations. Excessively high values combined with low confidence thresholds may result in multiple irregular instances being retained within a single physical field. |
| merge_iou | 0.8 | Minimum IoU required to merge instances originating from different tiles. Increasing this threshold may increase the number of seams in the final output, whereas decreasing it may cause excessive merging and produce unrealistically large polygon blocks. |
| merge_relative_area_threshold | 0.6 | Minimum overlap relative to the area of each instance required for merging. The condition must be satisfied for both instances. |
| merge_asymetric_pixel_area_threshold | 96 px | Minimum pixel overlap required for evaluating the merge_asymetric_relative_area_threshold condition. |
| merge_asymetric_relative_area_threshold | 0.7 | Minimum overlap relative to the area of an instance required to satisfy the asymmetric merging condition. |
| merging_edge_width | 4 px | Width of the region along the tile boundary in which alternative merging rules are applied. |
| merge_edge_iou | 0.6 | Minimum IoU between two instances within the tile-edge region required for merging. The IoU is calculated using only pixels within the edge region. |
| merge_edge_pixels | 192 px | Minimum number of overlapping pixels between two instances within the tile-edge region required for merging. |

Table S3. DelAnyFlow filtering parameters

| Parameter | Default value | Description |
|---|---|---|
| minimum_area_m2 | 2500 $m^2$ | Minimum area of an output polygon. The area of polygons evaluated using EPSG:6933. |
| minimum_part_area_m2 | 0 $m^2$ | Minimum area of polygon parts that can contribute to the construction of larger polygons. Increasing this value may result in small portions of fields being removed, particularly near region boundaries. |
| minimum_hole_area_m2 | 2500 $m^2$ | Minimum area of a hole that is retained within an output polygon. This value should not exceed minimum_area_m2, as doing so may result in spatial overlaps between output polygons. |

Table S4. DelAnyFlow simplification parameters

| Parameter | Default value | Description |
|---|---|---|
| epsilon_scale | 2 | Controls the degree of polygon simplification. Lower values preserve more detail and produce more geometrically complex boundaries, whereas higher values result in stronger simplification and smoother polygon boundaries. Directly scales automatically determined RDP tolerance. |

Table S5. DelAnyFlow additional parameters

| Parameter | Default value | Description |
|---|---|---|
| pixel_offset | [-1, -1] px | Specifies the systematic spatial offset applied to predicted instances to align them with their correct geographic position. This parameter is used to compensate for systematic misalignment in model predictions. |
| filter_classes | | Specifies the land-cover/land-use (LCLU) classes used to identify instances with excessive overlap with the corresponding classes. Overlapping pixels are retained in the instance geometry. This parameter is used only when an LCLU map is provided. |
| clip_classes | | Specifies the LCLU classes whose pixels are removed from the predicted instances. This parameter is used only when an LCLU map is provided. |

## S2. Strict spatial holdout and the Slovak test subset

### S2.1. Composition of the test split

The FBIS-22M test split combines two settings. The AI4Boundaries test regions and the two Ukrainian holdout regions have no spatial intersection with the training data. The Slovak test tiles reproduce the operational use of the model for national registry (LPIS) updating, in which the same parcels are observed on new acquisitions: the Slovak Sentinel-2 test tiles share footprints with training tiles but come from other dates, and the Slovak PlanetScope test tiles come from scenes spatially disjoint from the PlanetScope training scenes.

### S2.2. Image differences on shared footprints

Every Slovak Sentinel-2 footprint is covered in the training set by acquisitions from many dates, and the test tiles come from other dates. For randomly selected test and training pairs with the same footprint, the median Pearson correlation of the pixel values is only 0.59, so the test tiles are different images of the same area rather than copies of training images (Figure S1).

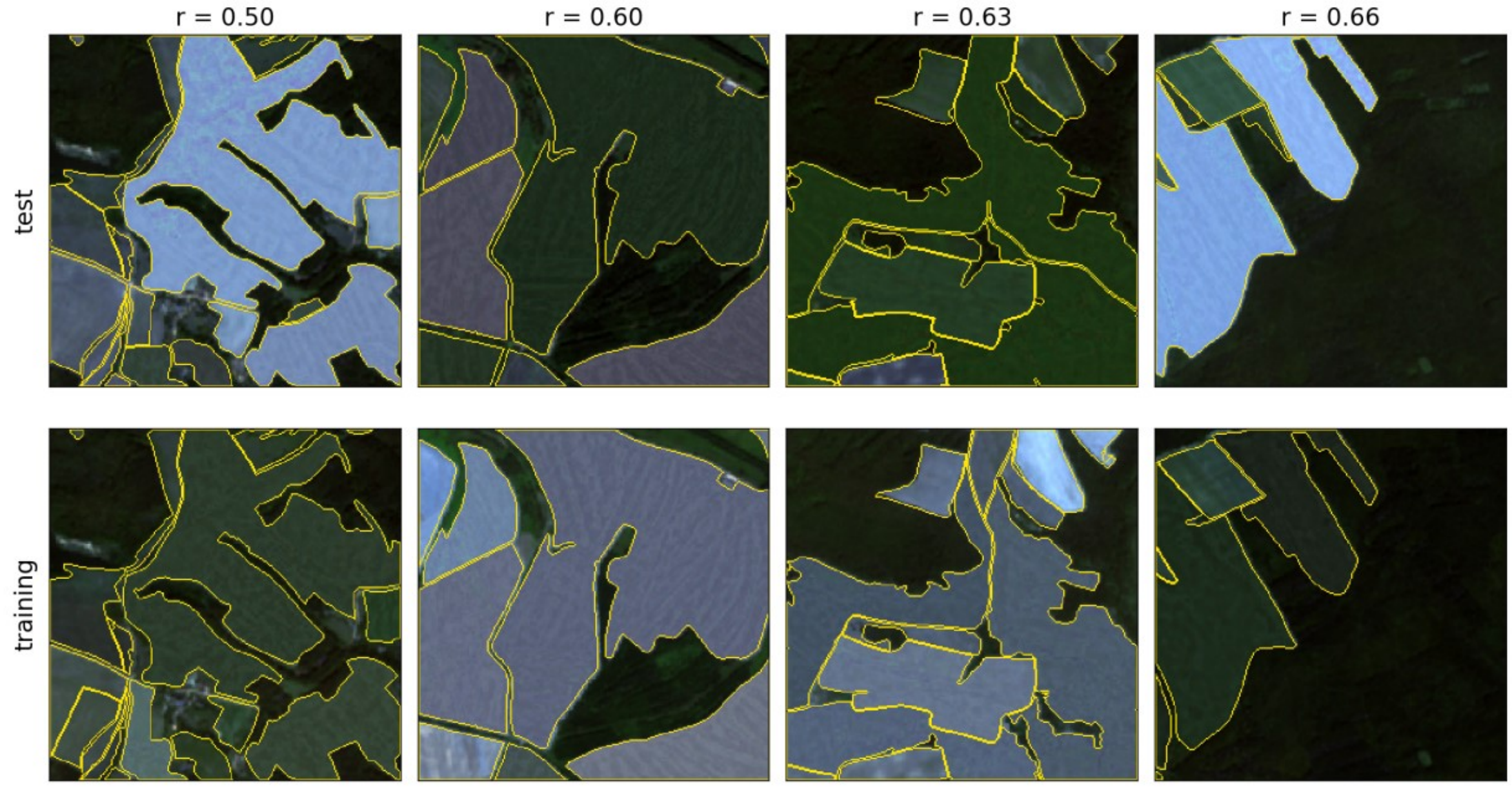

Figure S1. Slovak Sentinel-2 test tiles (top) and training tiles with the same footprint (bottom), with the reference boundaries (yellow). r is the Pearson correlation of the pixel values of each pair; the pairs are taken around the median of r.

### S2.3. Evaluation on the strict spatial holdout

All test tiles that overlap a training tile were removed, and DelAny and SAM2 were evaluated on the remaining tiles with the same protocol as in Table 2 of the main text (Table S6). Both models are less accurate on the strict holdout, including SAM2, which was never trained on any part of FBIS-22M. The Slovak parcels, taken from a carefully maintained registry, are therefore an easier part of the benchmark for every model, and the difference is not a consequence of the shared footprints alone. The ranking of the models and all conclusions of the comparison remain the same.

Table S6. Accuracy on the full test set and on the strict spatial holdout.

| Model | Test set | mAP@0.5 | mAP@0.5:0.95 |
|---|---|---|---|
| DelAny | Full test set (Table 2) | 0.720 | 0.477 |
| DelAny | Strict spatial holdout | 0.643 | 0.451 |
| SAM2 | Full test set (Table 2) | 0.382 | 0.235 |
| SAM2 | Strict spatial holdout | 0.246 | 0.178 |

## S3. Component-level ablation of cross-tile merging and anchor-guided simplification

### S3.1. Experimental setup

The national-scale experiment of the main text (Section 3.7) was repeated for the whole of Ukraine with the DelAny model and the default parameters of Tables S1 to S5, changing one component at a time. All variants process identical input imagery, so the differences between them are caused only by the post-processing. The metrics are computed on the complete maps and do not require reference boundaries.

### S3.2. Cross-tile merging

Three variants are compared: non-overlapping tiles without merging, overlapping tiles (tile step 0.5) without merging, and the full DelAnyFlow, with overlapping tiles and cross-tile merging. Duplication and fragmentation are measured as the number of polygons per km² of field area. A field is counted as cut by a seam when at least 40 m of its outline lies on a tile border, and two fields are counted as one field split in two when they share at least 40 m of a tile border. Because straight field boundaries can coincide with a border by chance, both counts are reported in excess of the same counts on a control grid shifted by half a tile. Omission and instance consistency near seams are assessed by comparing fields within 150 m of a tile border with fields in tile interiors, using the ratio of their median sizes and the field area near borders relative to the area expected from the share of the map covered by the border zone, where 1.0 means no loss.

Without merging, overlapping tiles produce almost twice as many polygons for the same field area, because every field covered by two tiles is returned as several fragments, and half of all fields are cut by a tile border (Table S7). Cross-tile merging reduces the number of fields split by a seam by a factor of four to six. Fields near tile borders become as large as, and cover as much area as, fields in tile interiors, whereas without merging they are about half the size and about 40% of the field area near borders is lost.

Table S7. Effect of tile overlap and cross-tile merging (whole of Ukraine).

| Variant | Polygons per km² of fields* | Fields cut by a seam (%)† | Split fields per 1,000† | Median size, seams / interior | Field area near seams |
|---|---|---|---|---|---|
| No overlap, no merging | 1.09 | 41.1 | 248 | 0.49 | 0.53 |
| Overlap, no merging | 1.90 | 52.5 | 358 | 0.46 | 0.61 |
| DelAnyFlow | 1.00 | 15.6 | 58 | 1.13 | 1.03 |

** relative to DelAnyFlow. † in excess of a control grid shifted by half a tile.*

### S3.3. Anchor-guided simplification

The raw polygons of the full DelAnyFlow run were simplified with the Ramer-Douglas-Peucker algorithm applied to each polygon independently, with the same algorithm in its topology-preserving form (GEOS), and with the anchor-guided simplification of DelAnyFlow. The tolerance of the two standard variants was set so that they keep the same number of vertices as the anchor-guided method, so that all methods compress the data equally. Overlaps are counted between pairs of fields whose simplified outlines intersect, and gaps as pairs of fields that share a boundary before simplification but no longer touch after it.

At the same compression, about 6% of the raw vertices, simplifying every polygon independently produces more than one overlapping pair per field and separates one in five pairs of neighbouring fields, because the two copies of each shared boundary are simplified differently (Table S8). The topology-preserving option keeps every polygon valid but does not keep shared boundaries identical, and gives the same result. The anchor-guided simplification fixes the vertices where shared boundaries begin and end, so that both neighbours receive the same simplified boundary, and removes the overlaps and gaps.

Table S8. Topological consistency after simplification at equal vertex count (whole of Ukraine).

| Method | Vertices per field | Overlapping pairs per 1,000 fields | Overlap area (% of fields) | Neighbours separated by a gap (%) |
|---|---|---|---|---|
| Raw outlines | 167.7 | 0 | 0 | 0 |
| RDP per polygon | 10.5 | 1,112 | 0.44 | 20.2 |
| RDP per polygon, topology-preserving | 10.5 | 1,119 | 0.45 | 19.9 |
| Anchor-guided RDP (DelAnyFlow) | 10.5 | <0.01 | 0.00 | 0.2 |

## S4. Delineation from inputs other than optical RGB imagery

DelAny takes a three-band image as input, so any data source that can be rendered as a three-channel image can be passed to it without changing the model. To illustrate this, DelAny was applied without retraining or fine-tuning to three inputs that are not native optical RGB imagery: a harmonic composite derived from a Sentinel-2 NDVI time series (Figure S2), a hue-shifted Sentinel-1 SAR composite (Figure S3), and a principal component projection of learned Earth observation embeddings (Figure S4). These examples are qualitative and cover limited areas. They demonstrate feasibility and do not establish an accuracy comparable to that obtained with optical inputs.

### S4.1. Harmonic NDVI composite

Following Papić et al. (2025), who recoloured the harmonic NDVI to create segmentation cues, an NDVI time series was converted into a three-channel image by harmonic regression and used in place of the Sentinel-2 median composite of 1 June to 30 July, which is the default input in this study, including the national map of Ukraine. The time series consisted of five Sentinel-2 composites over Moldova in 2025: 20 March to 25 April, 1 May to 15 June, 1 June to 5 July, 1 to 30 July, and 1 to 15 August. The same type of input is evaluated quantitatively in the main text (Table 6). Figure S2 shows the resulting image and the delineation obtained from it.

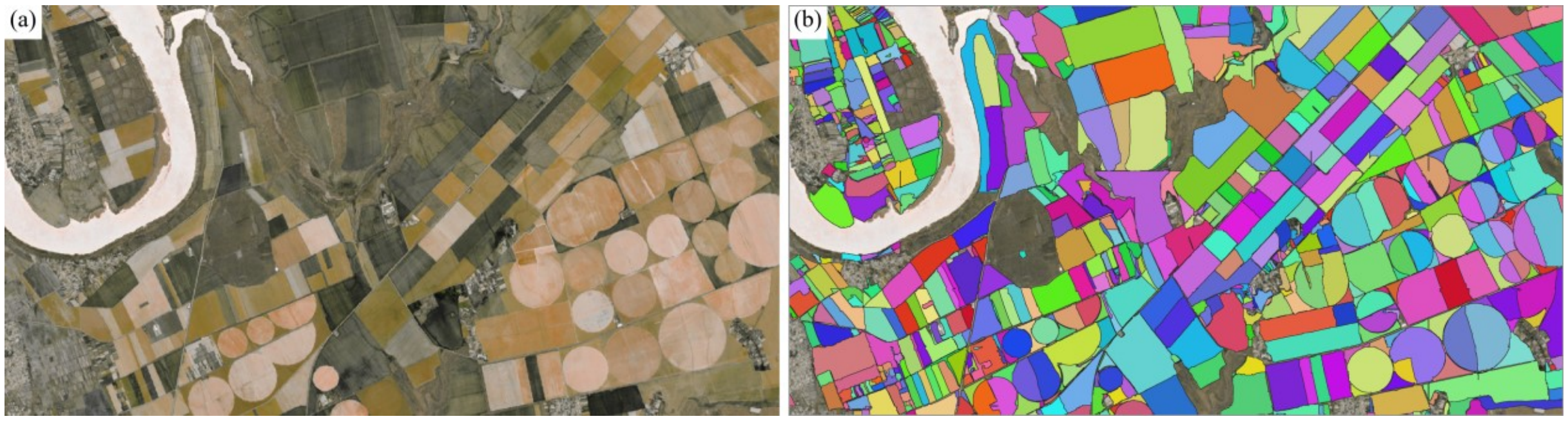

Figure S2. Delineation from a harmonic NDVI composite, Moldova, 2025 (47.3418° N, 29.2471° E): (a) three-channel image generated from a five-date Sentinel-2 NDVI time series; (b) delineation result.

### S4.2. Sentinel-1 SAR composite

Radar input was tested in Vinnytska oblast, Ukraine, with 2025 imagery. A three-channel composite was built from the mean Sentinel-1 VV backscatter in three successive monthly windows: 15 May to 15 June (red), 15 June to 15 July (green) and 15 July to 15 August (blue). To bring the composite closer to the colour distribution of optical vegetation imagery, a hue shift was applied: the composite was converted to HSV, the mean hue of the image patch was computed, all hue values were shifted by the difference between a target green hue and this mean, and the result was converted back to RGB. Figure S3 shows the composite and the resulting delineation.

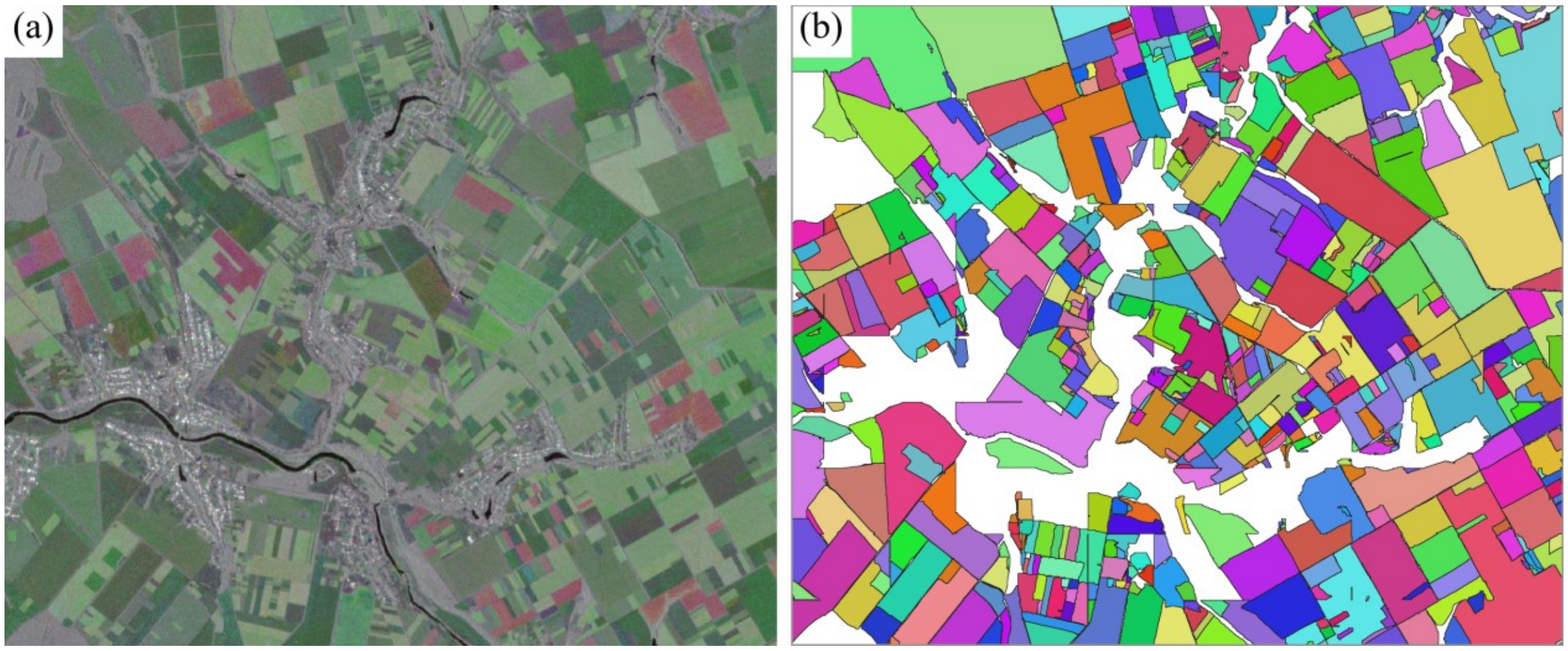

Figure S3. Delineation from a hue-shifted Sentinel-1 VV composite, Vinnytska oblast, Ukraine, 2025 (48.5174° N, 29.6842° E): (a) hue-shifted composite; (b) delineation result.

### S4.3. Learned Earth observation embeddings

Learned representations were tested with TESSERA v1.1 embeddings (Feng et al., 2025; https://github.com/ucam-eo/geotessera) for 2025 over Ternopilska oblast, Ukraine. TESSERA is a foundation model that encodes the annual Sentinel-1 and Sentinel-2 time series of each 10 m pixel into a 128-dimensional embedding, so the embedding image combines radar and optical information and has no direct visual interpretation. The 128-channel embedding image was reduced to three principal components (scikit-learn), which were used as the red, green and blue channels. Figure S4 shows the input and the delineation result.

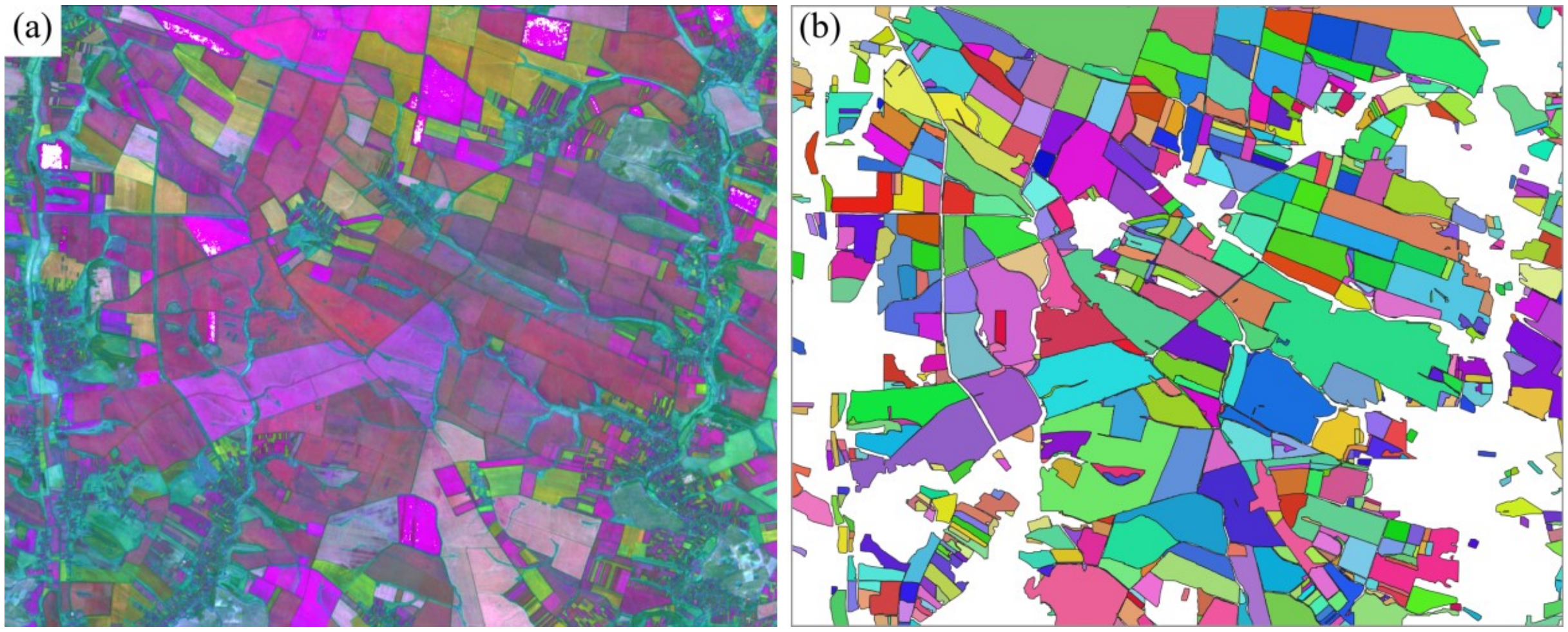

Figure S4. Delineation from a principal component projection of TESSERA embeddings, Ternopilska oblast, Ukraine, 2025 (49.2004° N, 25.2656° E): (a) input image; (b) delineation result.

## S5. Cross-region evaluation on an independent 100-country benchmark

FBIS-22M covers nine European countries (Section 2.2 of the main text). To quantify transfer to other regions, DelAny was evaluated on the independent, publicly available benchmark introduced by Lavreniuk et al. (2026). The benchmark contains four manually annotated image patches for each of 100 countries, so that every country has the same weight, and none of it was used for training. DelAny reaches an mAP@0.5 of 0.275 over all countries, with 0.332 in Europe, 0.317 in North America, 0.314 in Latin America, 0.251 in Africa, and 0.161 in Asia and Oceania. In North and Latin America, which are absent from FBIS-22M, the accuracy is almost the same as in Europe, the only region represented in training. The accuracy is lower in Africa and in Asia and Oceania, where small and fragmented smallholder fields with weak boundaries make delineation harder. Since none of these regions was seen in training, the results quantify the zero-shot transfer of DelAny: the model maintains its accuracy in new agricultural regions, and the differences between regions follow the complexity of the landscape rather than its presence in the training data.